\documentclass[lettersize,journal]{IEEEtran}
\usepackage{amsmath,amsfonts}
\usepackage{algorithmic}
\usepackage{algorithm}
\usepackage{array}
\usepackage[caption=false,font=normalsize,labelfont=sf,textfont=sf]{subfig}
\usepackage{textcomp}
\usepackage{stfloats}
\usepackage{url}
\usepackage{verbatim}
\usepackage{graphicx}
\usepackage{cite}
\usepackage{tikz}
\usepackage{comment}
\usepackage{amsmath,amssymb} 
\usepackage{color}
\usepackage{graphicx}
\usepackage{booktabs}
\usepackage{bbm}
\usepackage{makecell}
\usepackage{bm}
\usepackage{multicol}
\usepackage{multirow}
\usepackage{caption}
\usepackage{xcolor}
\usepackage{bbding}
\usepackage{pifont}
\usepackage{wrapfig}

\newtheorem{proposition}{Proposition}

\newcommand{\etc}{\textit{etc}}
\newcommand{\ie}{\textit{i.e.}}
\newcommand{\eg}{\textit{e.g.}}
\newcommand{\etal}{\textit{et al. }}

\long\def\comment#1{}

\begin{document}

\title{Imperceptible and Robust Backdoor Attack in\\ 3D Point Cloud}

\author{Kuofeng Gao \textsuperscript{*}\thanks{* Equal contribution.}, Jiawang Bai \textsuperscript{*}, Baoyuan Wu, \textit{Member IEEE},\\ Mengxi Ya, Shu-Tao Xia, \textit{Member IEEE}
\thanks{
Kuofeng Gao, Jiawang Bai, Mengxi Ya and Shu-Tao Xia are with Tsinghua Shenzhen International Graduate School, Tsinghua University, Shenzhen, Guangdong 518055, China and Shu-Tao Xia is also with the Peng Cheng Laboratory, Shenzhen, Guangdong 518055, China.
(E-mail: gkf21@mails.tsinghua.edu.cn, bjw19@mails.tsinghua.edu.cn, yamx21@mails.tsinghua.edu.cn, xiast@sz.tsinghua.edu.cn).}
\thanks{
Baoyuan Wu is with School of Data Science, the Chinese University of
Hong Kong, Shenzhen (CUHK-Shenzhen) and Secure Computing Lab of Big
Data, Shenzhen Research Institute of Big Data (SBRID), Shenzhen 518172,
China. (E-mail: wubaoyuan@cuhk.edu.cn)
}
\thanks{Corresponding authors: Baoyuan Wu. (E-mail: wubaoyuan@cuhk.edu.cn) and Shu-Tao Xia (E-mail: xiast@sz.tsinghua.edu.cn).}

}

\markboth{Gao and Bai \MakeLowercase{\textit{et al.}}: Imperceptible and Robust Backdoor Attack in 3D Point Cloud}%
{Shell \MakeLowercase{\textit{et al.}}: A Sample Article Using IEEEtran.cls for IEEE Journals}


\maketitle

\begin{abstract}
With the thriving of deep learning in processing point cloud data, recent works show that backdoor attacks pose a severe security threat to 3D vision applications. The attacker injects the backdoor into the 3D model by poisoning a few training samples with trigger, such that the backdoored model performs well on clean samples but behaves maliciously when the trigger pattern appears. 
Existing attacks often insert some additional points into the point cloud as the trigger, or utilize a linear transformation (\eg, rotation) to construct the poisoned point cloud. 
However, the effects of these poisoned samples are likely to be weakened or even eliminated by some commonly used pre-processing techniques for 3D point cloud, \eg, outlier removal or rotation augmentation.
In this paper, we propose a novel \textbf{i}mperceptible and \textbf{r}obust \textbf{b}ackdoor \textbf{a}ttack (IRBA) to tackle this challenge. 
We utilize a nonlinear and local transformation, called \textit{weighted local transformation} (WLT), to construct poisoned samples with unique transformations. 
As there are several hyper-parameters and randomness in WLT, it is difficult to produce two similar transformations. Consequently, poisoned samples with unique transformations are likely to be resistant to aforementioned pre-processing techniques. 
Besides, as the controllability and smoothness of the distortion caused by a fixed WLT, the generated poisoned samples are also imperceptible to human inspection.
Extensive experiments on three benchmark datasets and four models show that IRBA achieves $80\%+$ ASR in most cases even with pre-processing techniques, which is significantly higher than previous state-of-the-art attacks.
\end{abstract}


\section{Introduction}
\label{sec:introduction}
\IEEEPARstart{V}{ision} for 3D has been developed rapidly and become more popular in real-world applications, such as autonomous driving \cite{chen2017multi,li2020deep}, robot industry \cite{guo2019local}, and augmented reality \cite{mildenhall2020nerf,blanc2020genuage}, \etc. Since PointNet \cite{qi2017pointnet} was first proposed, more deep learning-based methods have been introduced into 3D domain and shown tremendous success in various tasks, \eg, point cloud classification. Although large progress has been made, the security problems of 3D deep learning are not explored systematically. Recently, 3D deep learning has been uncovered vulnerable to backdoor attacks, a training-time attack paradigm via data poisoning \cite{xiang2021backdoor,li2021pointba,huang2022backdoor,wu2022backdoorbench,zheng2022data}. The model trained on the poisoned dataset will classify to an attacker-specified label maliciously when the samples with the trigger present otherwise behave normally. Once the backdoored model \cite{nguyen2020input,doan2021backdoor,li2020backdoor,wang2020attack,bai2021targeted,xiang2022detecting} deploys in the safety-critical scenarios, it may lead to serious disasters.\par

\begin{figure*}[t]
    \begin{minipage}{\linewidth}
    \centering
    \includegraphics[width=0.85\linewidth]{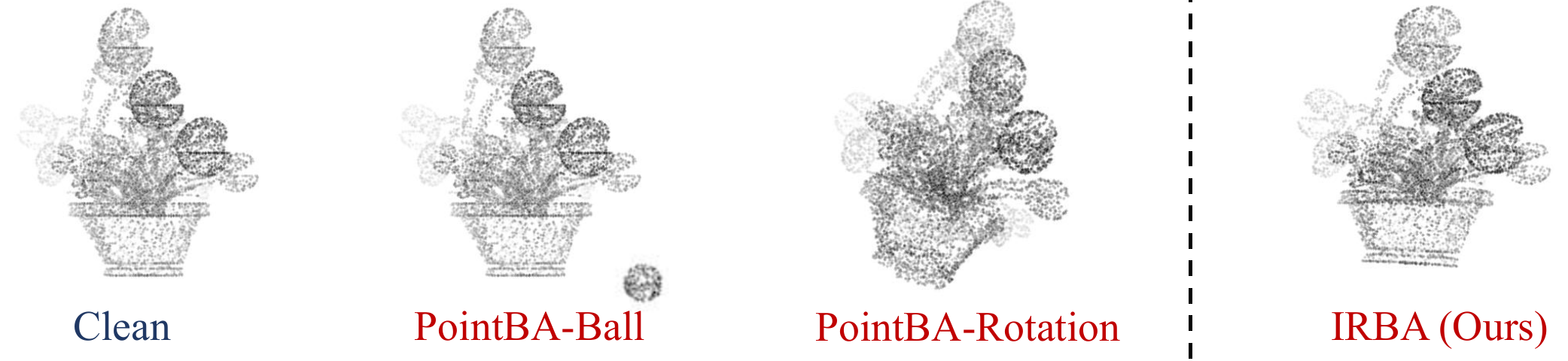}
    \end{minipage}
    \caption{Visualization of the clean point cloud and poisoned point clouds from different backdoor attacks, including the PointBA-Ball attack \cite{xiang2021backdoor, li2021pointba}, the PointBA-Rotation attack \cite{li2021pointba}, and our IRBA.}
	\label{fig:example}
\end{figure*}

In existing backdoor attacks for 3D deep learning, the poisoned samples are constructed by inserting some additional points (\eg, the small ball shown in the second column of Fig. \ref{fig:example}) \cite{xiang2021backdoor,li2021pointba,tian2021poisoning}, or rotating the original point cloud \cite{li2021pointba} (\eg, the third column of Fig. \ref{fig:example}). However, these attacks may not succeed in practice, as their poisoned samples are not resistant to some pre-processing techniques that have been commonly applied to pre-process point clouds during model training \cite{qi2017pointnet,qi2017pointnet++}. 
For example, the trigger via additional points can be easily removed by statistical outlier removal (SOR) \cite{zhou2019dup}, and the effect of the rotated poisoned samples can be mitigated by the rotation augmentation. 
Besides, the trigger via additional points is noticeable to humans.
Hence, a practical backdoor attack against point cloud models should be not only robust to aforementioned pre-processing techniques, but also imperceptible to human inspection.


In this work, we focus on the transformation-based approach to construct poisoned samples, as the additional point-based approach is perceptible and cannot surpass the SOR operation. 
However, as mentioned above, the transformation-based poisoned samples may be vulnerable to the data augmentation adopted in model training. 
One possible reason we speculate is that the data augmentation could produce augmented samples with similar transformations with poisoned samples, such that the steady mapping between poisoned samples and the target class cannot be learned by the model. 
To verify this point, we conduct a brief experiment by constructing poisoned samples with a 10$^\circ$ rotation, and adopt the random rotation augmentation in model training, with the range of $(0^\circ, 10^\circ), (0^\circ, 20^\circ), (0^\circ, 360^\circ)$, respectively. The corresponding attack success rates (ASR) are 97.5\%, 11.7\% and 3.4\%, respectively. It tells that if the rotation used for poisoned samples is covered by the random rotation augmentation, then the attack performance dramatically drops, which verifies our speculation to some extent.

Thus, we believe more complex transformations that are difficult to reproduce are desired to construct poisoned samples. 
We choose the \textit{weighted local transformation} (WLT) \cite{kim2021point}, which was proposed as an augmentation technique to learn better 3D point cloud models.
As shown in Fig. \ref{fig:pipeline}-(a), it consists of two steps: firstly conducting global transformations (\eg, rotation, scaling) around multiple randomly used anchor points to obtain multiple transformed samples, then merging these samples into one unified transformed sample through smooth aggregation. Note that the distortion between the final transformed sample and the original sample is nonlinear and local (\ie, the distortion is local location). 
It is determined by several hyper-parameters (including the number of anchors, the types and values of global transformations), and the randomness on anchor sampling and specific global transformation. 
Even with the same hyper-parameters, it is also difficult to obtain two similar transformed samples if conducting WLT transformations twice. 
The reproducing difficulty facilitates to produce unique transformations. 

Inspired by the above analysis, we propose an effective backdoor attack, called \textit{\textbf{i}mperceptible and \textbf{r}obust \textbf{b}ackdoor \textbf{a}ttack} (IRBA). It utilizes the \textit{WLT with fully fixed hyper-parameters} (\ie, fixed anchors, fixed global transformations) to generate poisoned samples with a unique nonlinear and local transformation.
Even the defender knows the WLT is used for poisoning and also utilizes WLT for data augmentation in training, it is still difficult to generate augmented samples with similar transformations compared to poisoned samples.  
Consequently, the WLT-based poisoned samples are resistant to the data augmentations (no matter WLT or any global transformation) adopted in model training, which will be experimentally verified later (see Section \ref{sec: adpative defense} and Table \ref{fig:trigger parameter}). 
Moreover, due to the controllability (\ie, the distortion of every point could be exactly computed) and smoothness of the distortion, WLT-based poisoned samples are also imperceptible to human inspection and can bypass the SOR operation (see the last column of Fig. \ref{fig:example}), which will be analyzed and verified later.

In summary, our main contributions are three-fold.\par 
\begin{itemize}
    \item We demonstrate that existing backdoor attacks with PointBA-Rotation poisoned samples are vulnerable to random data augmentation and provide a reasonable explanation.
    \item We propose an imperceptible and robust backdoor attack method by utilizing a nonlinear and local transformation, weighted local transformation (WLT).
    \item We conduct extensive experiments to show the superiority of IRBA to previous state-of-the-art backdoor attacks in 3D point cloud.
\end{itemize}

\comment{
\begin{itemize}
\item We develop the imperceptible and robust backdoor attack (IRBA) in 3D point cloud to overcome the vulnerability of existing backdoor attacks in practical scenarios.
\item We propose to utilize nonlinear operation to craft the poisoned point clouds, consisting of multi-anchor transformation and smooth aggregation. The poisoned samples are proven to be robust to pre-processing techniques and imperceptible to human inspection. 
\item We conduct extensive experiments to show the superiority of IRBA. Especially, under SOR and rotation augmentation, the success rate of IRBA is greater than 80\% in most cases, compared to less than 30\% success rate of previous attacks.
\end{itemize}
}

\section{Related work}
\subsection{Deep learning in 3D point cloud} 
Point clouds are a well-known data structure in 3D domain, which contain a set of unordered point coordinates. Promoted by the deep learning, point-based learning \cite{zhao2019pointweb,simonovsky2017dynamic,xu2018spidercnn,mao2019interpolated,qin2019pointdan} has become increasingly popular due to its promising performance. Qi \etal \cite{qi2017pointnet} first designed a novel neural network PointNet to directly learn the point-wise features. It adopts the symmetric function, max-pooling, to preserve the order-invariant property of point clouds. Motivated by PointNet, PointNet++ \cite{qi2017pointnet++} further exploits the set abstraction layers to enhance the multi-scale information extraction. Moreover, the graph neural network and X-convolution operation are introduced in DGCNN \cite{wang2019dynamic} and PointCNN \cite{li2018pointcnn} to obtain more compact representations. The above architectures have been extended to other complex tasks in 3D point cloud, such as 3D semantic segmentation \cite{landrieu2018large,zhang2020deep} and 3D object detection \cite{zhou2018voxelnet,lang2019pointpillars,wang2021object},
but the security problems like backdoor attacks have not been explored well. 


\subsection{Backdoor attack in 3D point cloud} 
Li \etal \cite{li2021pointba} first extended the backdoor attack to 3D point cloud. Analog with stamping a patch at the corner of an image in 2D backdoor attack \cite{gu2017badnets}, they launched additional points like a ball near the 3D object to inject the 3D deep learning. Xiang \etal \cite{xiang2021backdoor} generated backdoor points by optimizing their spatial locations and performing the attack under a strict setting without knowing the victim architecture successfully. However, these two ball-based attacks can be discovered easily by humans. To bypass the human inspection, Li \etal \cite{li2021pointba} also investigated the rotation operation as the 3D backdoor trigger and implement an imperceptible attack.\par


The previous works \cite{xiang2021backdoor,li2021pointba} in 3D backdoor learning provide two potential mitigation methods customized for 3D backdoor attacks. One is the statistical outlier removal which uses the \textit{k}-nearest neighbors (kNN) algorithm to define the distance metric to remove the abnormal points in a point cloud. Therefore, the attack performance of the PointBA-Ball attack can be easily mitigated by SOR. The other is inspired by the common 3D point cloud augmentation technique, including rotation, scaling, and point-wise jitter, \etc. Since 3D deep learning usually adopts them to improve the generalization of the model, the PointBA-Rotation attack will be erased during the actual training. 

\section{Threat model}

We consider a classical scenario following existing works \cite{gu2017badnets,chen2017targeted,li2021invisible} , poison-label backdoor attack, where the attacker is allowed to craft a small number of the poisoned samples and poison them into the training set. But the attacker can not know the victim architecture or control over the backdoor training process. Therefore, in the realistic attack scenario, the poisoned training samples may be pre-processed before they are fed into the model \cite{li2021pointba,xiang2021backdoor,borgnia2021strong,zeng2020deepsweep}.

The goal of the attacker is that any model trained on the poisoned dataset will return the attacker-specified malicious target label when meeting the point clouds injected with the trigger pattern. Meanwhile, it should also preserve the clean accuracy in the absence of the trigger pattern. In addition to the effectiveness of the backdoor attack, the generation of poisoned samples should not destroy the geometric shape of the point clouds significantly to ensure the imperceptibility. 

\begin{figure*}[t]
    \centering
    \includegraphics[width=\linewidth]{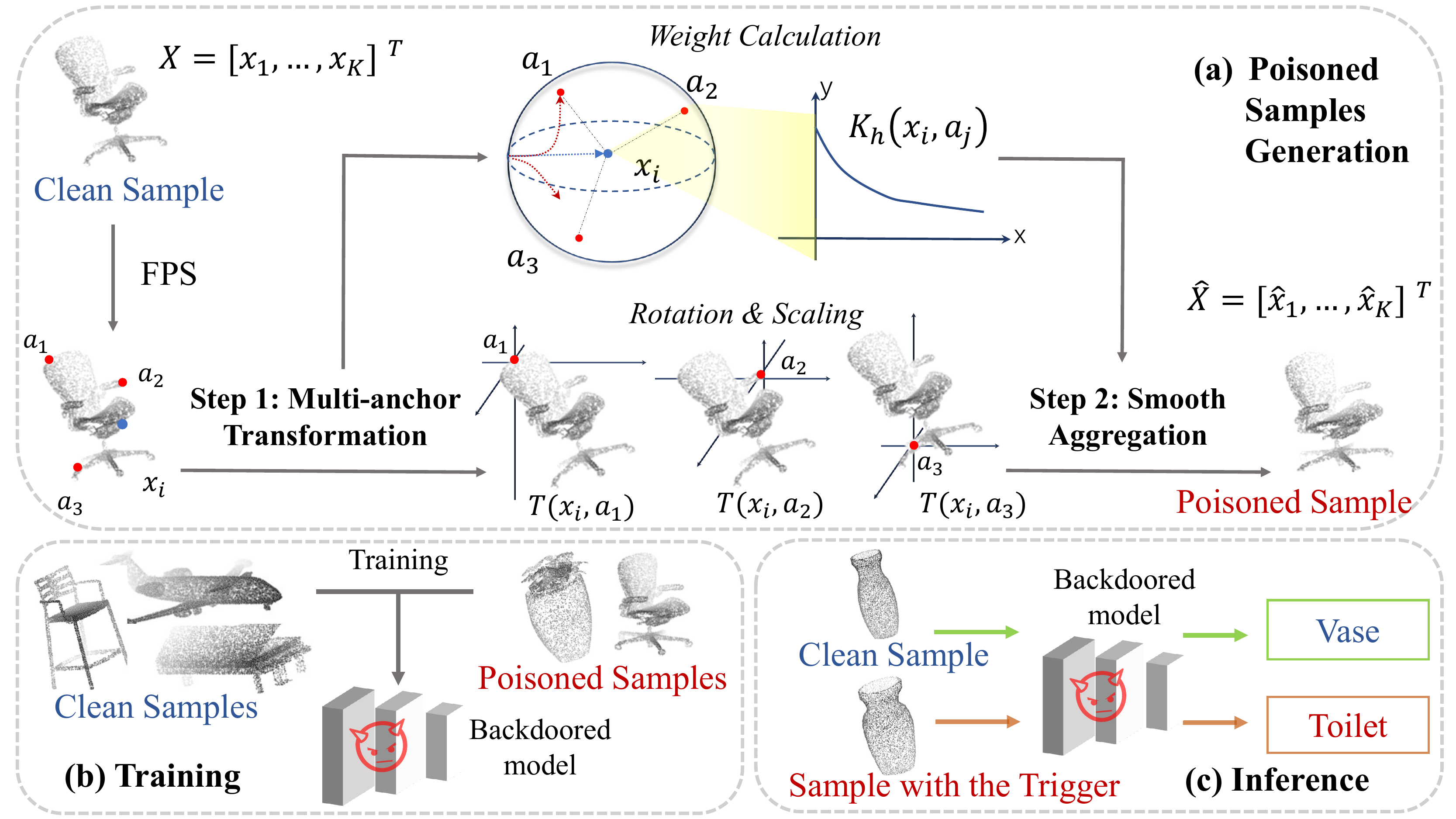}
    \caption{The pipeline of our proposed IRBA. (a) Generating poisoned samples by WLT, including multi-anchor transformation and smooth aggregation. (b) Training on the poisoned dataset which contains both clean samples and poisoned samples. (c) Inference by the backdoored model. }
	\label{fig:pipeline}
\end{figure*}

\section{Imperceptible and robust backdoor attack (IRBA)}

\subsection{Problem definition}
Given a training set $\mathcal{D}=\{(\bm{X}_i,y_i)\}^N_{i=1}$ which contains $N$ point cloud samples, where $y \in \{1,2,...,C\}$ denotes the ground-truth label of the point cloud sample and $C$ is the number of class. $\bm{X}_i=[\bm{x}_1, \bm{x}_2, ..., \bm{x}_K]^\top \in \mathbb{R}^{K \times 3}$ denotes that every point cloud can be decomposed into $K$ points and each point $\bm{x}_i \in \mathbb{R}^{3}$ has its $3$-dimension position coordinate. 

Based on original $M$ point cloud samples, the attacker crafts a small number of the poisoned samples as $\mathcal{D}_{b}=\{(\hat{\bm{X}}_i,y_t)^{(i)}\}^M_{i=1}$, where $y_t$ is the target label and $M\ll N$. $\mathcal{D}_{c}=\{(\bm{X}_i,y_i)\}^{N-M}_{i=1}$ denotes the remain clean sample set. The poisoned dataset is built by mixing $\mathcal{D}_{b}$ and $\mathcal{D}_{c}$. $f_{\bm{\theta}}$ denotes the victim classification model with parameters $\bm{\theta}$. Because we assume that the attacker can not change the training process, $f_{\bm{\theta}}$ should be obtained by normally training on the poisoned dataset with minimizing the following objective:
\begin{equation}
\begin{aligned}
\min_{\bm{\theta}} \sum_{(\bm{X}, y) \in \mathcal{D}_{c} \cup \mathcal{D}_{b}} \mathcal{L}\left(f_{\bm{\theta}}(\mathcal{P}(\bm{X})), y\right),
\end{aligned}
\label{eq:backdoor training}
\end{equation}
where $\mathcal{L}(\cdot,\cdot)$ indicates the loss function during the training stage. We also consider the pre-processing on the training samples, denoted as $\mathcal{P}(\cdot)$, including SOR or 3D data augmentations. They have become common configurations for cleaning the point clouds or improving the performance of the 3D model \cite{qi2017pointnet,qi2017pointnet++,zhou2019dup}. Hence, it is necessary to ensure the backdoor can be injected even the poisoned samples are processed by $\mathcal{P}(\cdot)$.



\subsection{WLT-based poisoned samples generation}

The main focus of this paper is to craft poisoned samples which are robust to various pre-processing techniques. Different from inserting additional points in \cite{xiang2021backdoor, li2021pointba} and the linear transformation in \cite{li2021pointba}, we utilize WLT to generate each point $\hat{\bm{x}}_i$ in the poisoned sample $\hat{\bm{X}}$ by a nonlinear point-wise function $G: \mathbb{R}^{3} \rightarrow \mathbb{R}^{3}$. The pipeline is described in Fig. \ref{fig:pipeline}.

\noindent \textbf{Multi-anchor transformation.} We firstly select the anchor points before performing the transformation.  In order to ensure the anchors don't gather in one local region to cause the uneven deformation, we utilize Farthest Point Sampling (FPS) algorithm to choose a few transformation anchors $\{\bm{a}_j\}^W_{j=1}$ on the surface of the point cloud $\bm{X}$, where $W$ is the number of anchor points.  FPS first samples one initial point randomly and repeatedly selects the points farthest from the previous points. It can lead to more subtle variations on the 3D shape and meanwhile introduce more randomness.

After selecting the anchor points, we perform anchor-based transformation by taking each anchor point as the centroid and applying the rotation and scaling operation. We implement the transformation by a rotation matrix $\bm{R} \in \mathbb{R}^{3 \times 3}$ and a scaling matrix $\bm{S} \in \mathbb{R}^{3 \times 3}$. For simplicity, we rotate the 3D point cloud along the $x$, $y$, and $z$-axis with the same angle $\alpha$. We define the rotation matrix in multi-anchor transformation as $\bm{R}=\bm{R}_{x}(\alpha)\bm{R}_{y}(\alpha)\bm{R}_{z}(\alpha)$, which represents the basic 3D point cloud rotation by an angle $\alpha$ along three axes in order. The formulations of $\bm{R}_{x}(\alpha)$, $\bm{R}_{y}(\alpha)$, and
$\bm{R}_{z}(\alpha)$ are as follows:
\begin{equation}
\begin{aligned}
&\bm{R}_{x}(\alpha)=\left[\begin{array}{ccc}
1 & 0 & 0 \\
0 & \cos \alpha & -\sin \alpha \\
0 & \sin \alpha & \cos \alpha
\end{array}\right], \\
&\bm{R}_{y}(\alpha)=\left[\begin{array}{ccc}
\cos \alpha & 0 & \sin \alpha \\
0 & 1 & 0 \\
-\sin \alpha & 0 & \cos \alpha
\end{array}\right], \\
&\bm{R}_{z}(\alpha)=\left[\begin{array}{ccc}
\cos \alpha & -\sin \alpha & 0 \\
\sin \alpha & \cos \alpha & 0 \\
0 & 0 & 1
\end{array}\right],
\end{aligned}
\end{equation} 
The scaling matrix is defined as a diagonal matrix, denoted as $\bm{S}=\text{diag}(s, s, s)$, which means scaling with the same size $s$ along three axes. Based on the selected anchor $\bm{a}_j$ and the rotation and scaling operation, we define the anchor-based transformation for an input point $\bm{x}_i$ as follows:
\begin{equation}
\begin{aligned}
    T(\bm{x}_i, \bm{a}_j )=\bm{R}\bm{S}(\bm{x}_i-\bm{a}_{j})+\bm{a}_{j}. 
\end{aligned}
\label{eq:local transformation}
\end{equation}
Note that the rotation matrix and scaling matrix are predefined by the attacker and fixed to generate all poisoned samples.
For each anchor point $\bm{a}_j$, we perform the above transformation for each point in $\bm{X}$, resulting in $W$ transformed point clouds.

\begin{figure*}[t]
    \begin{minipage}{\linewidth}
    \centering
    \includegraphics[width=0.9\linewidth]{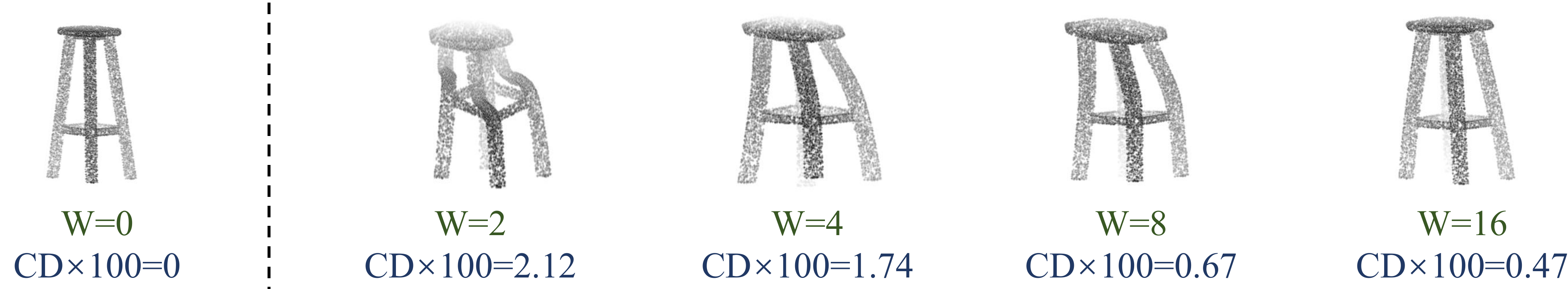}
    \end{minipage}
    \caption{Poisoned samples generated by WLT under different number of the anchors $W$. The CD score \cite{liu2020morphing,yuan2018pcn} is calculated between the clean sample and the sample with the trigger, where the lower CD score corresponds to more imperceptibility.}
	\label{fig:vis_w_parameter}
\end{figure*}
\begin{figure}[t]
    \includegraphics[width=\linewidth]{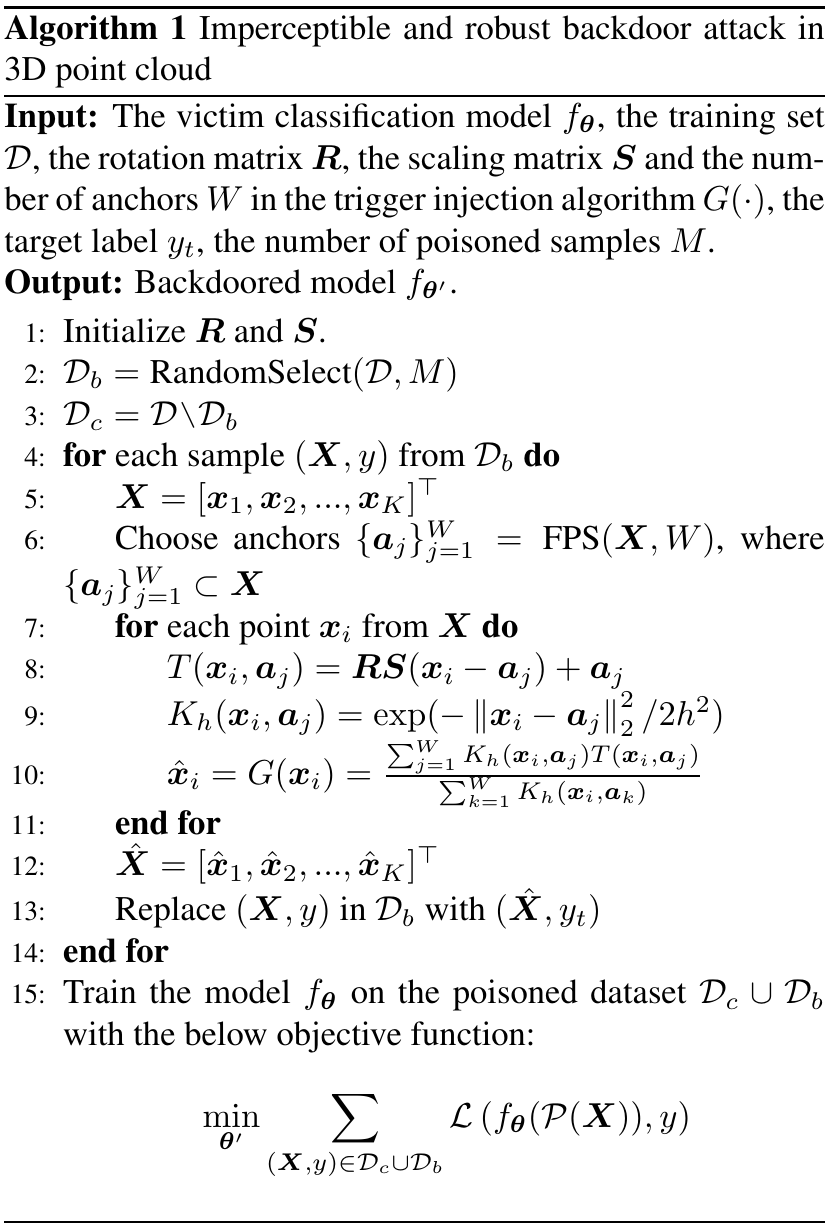}
\end{figure}

\noindent \textbf{Smooth aggregation.} In the step of aggregating the $W$ transformed point clouds, we introduce the Gaussian kernel function $K_{h}(\bm{x}_i, \bm{a}_{j})=\exp (-\left\|\bm{x}_i- \bm{a}_{j}\right\|_{2}^{2}/ 2 h^{2})$ to calculate the weight for each pair of $\bm{x}_i$ and $\bm{a}_{j}$, where the bandwidth $h$ is a smoothing parameter. Given the kernel function, the point $\hat{\bm{x}}_i$ in the poisoned sample $\hat{X}$ can be generated as:
\begin{equation}
\begin{aligned}
    \hat{\bm{x}}_i=G(\bm{x}_i)=\frac{\sum_{j=1}^{W} K_{h}\left(\bm{x}_i, \bm{a}_{j}\right) T(\bm{x}_i, \bm{a}_j)}{\sum_{k=1}^{W} K_{h}\left(\bm{x}_i, \bm{a}_{k}\right)}. 
\end{aligned}
\label{eq:smooth aggregation}
\end{equation}

\begin{proposition}
\label{proposition}
\textit{ (Originally defined in \cite{kim2021point}) Based on the Gaussian kernel, the aggregation function $G(\cdot)$ is smooth when  $T(\cdot, \cdot)$ is defined as Eq. (\ref{eq:local transformation}), where the smoothness of a function is that all partial derivatives exist and are continuous.}
\end{proposition}

One advantage of our aggregation is that $G(\cdot)$ introduces the nonlinear weighting due to the Gaussian kernel. Since common 3D data augmentations are linear, nonlinearity helps to improve the robustness of our attack when the poisoned samples are processed by $\mathcal{P}(\cdot)$. The other advantage is that our aggregation function is smooth according to Proposition \ref{proposition}. Therefore, the aggregation is prone to preserve the global shape of the point cloud and change its local structure in a smooth manner, which ensures the imperceptibility of the poisoned samples.  We visualize the poisoned samples in Fig. \ref{fig:vis_w_parameter}, which confirms the effect of our smooth aggregation. Besides, with the increasing of the number of anchors, the changes of the local structure become more imperceptible.



\begin{table*}[t]
	\centering
	\caption{ACC (\%) and ASR (\%) of backdoored models with the PointBA-Ball trigger, the PointBA-Rotation trigger and our IRBA trigger on three datasets. Our proposed IRBA is highlighted in \bf{bold}.}
	\resizebox{\linewidth}{!}{
	\begin{tabular}{l|cccccc|cccccc|cccccc}
		\hline
		\multicolumn{1}{l|}{\multirow{3}{*}{Models}} & \multicolumn{6}{c|}{ModelNet10} & \multicolumn{6}{c|}{ModelNet40} & \multicolumn{6}{c}{ShapeNetPart}  \\  \cline{2-19}
		& \multicolumn{2}{c}{PointBA-Ball} & \multicolumn{2}{c}{PointBA-Rotation} & \multicolumn{2}{c|}{\bf{IRBA (Ours)}} & \multicolumn{2}{c}{PointBA-Ball} & \multicolumn{2}{c}{PointBA-Rotation} & \multicolumn{2}{c|}{\bf{IRBA (Ours)}} & \multicolumn{2}{c}{PointBA-Ball} & \multicolumn{2}{c}{PointBA-Rotation} & \multicolumn{2}{c}{\bf{IRBA (Ours)}} \\ 
		& ACC & \multicolumn{1}{c}{ASR} & ACC & \multicolumn{1}{c}{ASR} & ACC & \multicolumn{1}{c|}{ASR} & ACC & \multicolumn{1}{c}{ASR} & ACC & \multicolumn{1}{c}{ASR} & ACC & \multicolumn{1}{c|}{ASR} & ACC & \multicolumn{1}{c}{ASR} & ACC & \multicolumn{1}{c}{ASR} & ACC & ASR \\ \hline
		PN & 92.5 & 100 & 91.3 & 97.4 & 91.8 & 97.0 & 86.4 & 99.8 & 86.8 & 94.6 & 87.1 & 93.2 & 98.3 & 100 & 98.2 & 97.7 & 98.1 & 99.2 \\ 
		PN++ & 93.2 & 100 & 92.3 & 98.3 & 93.2 & 98.1 & 89.6 & 100 & 89.9 & 95.8 & 90.5 & 95.7 & 98.6 & 100 & 98.4 & 97.1 & 98.8 & 99.0 \\
		DGCNN & 93.1 & 100 & 92.2 & 98.3 & 92.9 & 98.6 & 91.3 & 100 & 91.2 & 95.9 & 91.3 & 94.0 & 98.9 & 100 & 98.8 & 96.6 & 98.9 & 98.4 \\
        PCNN & 91.7 & 100 & 89.2 & 91.4 & 89.8 & 90.5 & 85.1 & 99.4 & 83.1 & 93.4 & 83.9 & 94.6 & 97.1 & 99.4 & 95.5 & 96.2 & 97.5 & 95.6 \\
		\hline
	\end{tabular}}
	\label{ASR of different methods}
\end{table*}

\begin{table*}[t]
\centering
\makebox[0pt][c]{\parbox{\textwidth}{%
    \begin{minipage}[t]{0.49\hsize}\centering
    	\caption{ACC (\%) of clean-trained models on three datasets across all architectures.}
    	\resizebox{0.95\linewidth}{!}
    	{
    	\begin{tabular}{l|ccc}
    		\hline
    		Models & ModelNet10 & ModelNet40 & ShapeNetPart  \\ \cline{1-4}
    		PN & 93.8 & 89.1 & 98.5 \\
    		PN++ & 94.8 & 91.5 & 99.1   \\
    		DGCNN & 94.5 & 92.0 & 99.1  \\
            PCNN & 91.1 & 86.9 & 98.4  \\
    		\hline
    	\end{tabular}
    	}
    	\label{acc of clean model}
    \end{minipage}
    \hfill
    \begin{minipage}[t]{0.49\hsize}\centering
    	\caption{ACC (\%) and ASR (\%) of IRBA on three datasets with different target labels. 
    	}
    	\resizebox{\linewidth}{!}
    	{
 	\begin{tabular}{l|cccccc}
		\hline
		\multicolumn{1}{l|}{\multirow{2}{*}{Datasets}} & \multicolumn{2}{c}{$y_t=0$} & \multicolumn{2}{c}{$y_t=1$} & \multicolumn{2}{c}{$y_t=2$}  \\ 
		& ACC & ASR & ACC & ASR & ACC & ASR \\
        \hline
		ModelNet10 & 93.8 & 98.7 & 92.8 & 98.8 & 93.2 & 99.6  \\
		ModelNet40 & 90.9 & 96.0 & 90.5 & 95.3 & 90.8 & 94.6   \\
		ShapeNetPart & 98.7 & 98.8 & 99.0 & 98.7 & 98.9 & 99.0   \\
		\hline
	\end{tabular}}
    	\label{different target label}
    \end{minipage}
}}
\end{table*}

\section{Experiments}
\subsection{Evaluation setup}
\label{sec:setting}

\noindent \textbf{Datasets and target models.} We adopt three benchmark datasets in the 3D point cloud classification task, \ie, ModelNet10 \cite{wu20153d}, ModelNet40 \cite{wu20153d} and ShapeNetPart \cite{yi2016scalable}. Following \cite{li2021pointba}, we build training and test point clouds for each dataset. Besides, we follow \cite{qi2017pointnet} to uniformly sample 1,024 points from the surface of each dataset and normalize them into a unit ball. 
We evaluate all attack methods on four categories of popular point cloud classification models, \ie, PointNet \cite{qi2017pointnet}, PintNet++ \cite{qi2017pointnet++}, DGCNN \cite{wang2019dynamic}, and PointCNN \cite{li2018pointcnn}, denoted as ``PN'', ``PN++'', ``DGCNN'' and ``PCNN''.\par

\noindent \textbf{Baseline methods.} We compare IRBA with two existing backdoor attacks in 3D point cloud classification in \cite{li2021pointba}: PointBA-Ball backdoor attack and PointBA-rotation backdoor attack (dubbed ``PointBA-Ball'' and ``PointBA-Rotation'' respectively). For the PointBA-ball attack, we place a ball with the fixed radius 0.05 centered at (0.05, 0.05, 0.05) near the 3D object, where the ratio between the trigger points and total points is 0.01. 
Besides, we rotate the 3D objects along the $z$-axis with 10° to craft poisoned samples for the PointBA-Rotation attack. \par

\noindent \textbf{Attack settings.}
For all methods, we set the poisoned rate $M/N=0.1$, which indicates the proportion of the number of poisoned samples in the poisoned dataset. Specially, poisoned samples are sampled from the non-target class randomly and the target label on ModelNet10, ModelNet40 and ShapeNetPart is chosen as ``Tabel''  ($y_t=8$), ``Toilet'' ($y_t=35$) and ``Lamp'' ($y_t=8$) respectively. For the proposed IRBA, we set the number of anchors $W$ as 16. The angle of the rotation $\alpha$ and the scaling size $s$ in the multi-anchor transformation is set as 5° and 5, respectively. The bandwidth $h$ in the smooth aggregation is set to 0.5. We use the framework PyTorch \cite{paszke2019pytorch} to implement all the experiments. We adopt the Adam optimizer \cite{kingma2014adam} with the learning rate 0.001 to train all the models for 200 epochs and the training batch size is set to 32. All the above experiments are run on a NVIDIA RTX A5000 GPU. Note that the training schedule on the poisoned dataset is the same as that on the clean dataset.\par

\noindent \textbf{Evaluation metrics.}
To verify the performance of backdoor attacks, we adopt attack success rate (ASR) as the metric, which is the fraction of samples from non-target class with the trigger classified to the target label by the backdoored model. A higher ASR means a stronger attack. Besides, we also use the test accuracy (ACC) to measure the effect of the backdoor attack on the clean dataset. Following \cite{liu2020morphing,yuan2018pcn,xiang2019generating},  we adopt the Chamfer Distance (CD) to investigate the deformation magnitude of poisoned point clouds compared to clean samples.


\begin{table*}[t]
	\centering
	\caption{ACC (\%) and ASR (\%) of backdoored models with the PointBA-Ball trigger, the PointBA-Rotation trigger and our IRBA trigger on three datasets under SOR and random rotation during training. The best results are highlighted in \bf{bold}.}

	\resizebox{\linewidth}{!}{
	\begin{tabular}{l|cccccc|cccccc|cccccc}
		\hline
		\multicolumn{1}{l|}{\multirow{3}{*}{Models}} & \multicolumn{6}{c|}{ModelNet10} & \multicolumn{6}{c|}{ModelNet40} & \multicolumn{6}{c}{ShapeNetPart}  \\  \cline{2-19}
		& \multicolumn{2}{c}{PointBA-Ball} & \multicolumn{2}{c}{PointBA-Rotation} & \multicolumn{2}{c|}{\bf{IRBA (Ours)}} & \multicolumn{2}{c}{PointBA-Ball} & \multicolumn{2}{c}{PointBA-Rotation} & \multicolumn{2}{c|}{\bf{IRBA (Ours)}} & \multicolumn{2}{c}{PointBA-Ball} & \multicolumn{2}{c}{PointBA-Rotation} & \multicolumn{2}{c}{\bf{IRBA (Ours)}} \\ 
		& ACC & \multicolumn{1}{c}{ASR} & ACC & \multicolumn{1}{c}{ASR} & ACC & \multicolumn{1}{c|}{ASR} & ACC & \multicolumn{1}{c}{ASR} & ACC & \multicolumn{1}{c}{ASR} & ACC & \multicolumn{1}{c|}{ASR} & ACC & \multicolumn{1}{c}{ASR} & ACC & \multicolumn{1}{c}{ASR} & ACC & ASR \\ \hline
		PN &  84.2 & 19.4 & 86.9 & 8.17 & \textbf{88.5} & \textbf{90.6} & 78.7 & 14.2 & 82.1 & 9.97 & \textbf{85.7} & \textbf{83.7} & 93.6 & 21.4 & 97.6 & 14.5 & \textbf{97.7} & \textbf{98.8} \\ 
		PN++ & 83.1 & 11.7 & 88.8 & 8.41 & \textbf{92.4} & \textbf{96.6} & 78.4 & 19.7 & 82.5 & 11.4 & \textbf{88.1} & \textbf{90.2} & 87.8 & 29.6 & 94.3 & 10.7 & \textbf{98.5} & \textbf{98.8} \\
		DGCNN & 91.8 & 10.2 & 91.1 & 6.97 & \textbf{92.9} & \textbf{92.5} & 87.8 & 7.82 & 88.7 & 4.23 & \textbf{90.0} & \textbf{85.1} & 94.0 & 6.07 & 96.8 & 7.76 & \textbf{98.4} & \textbf{97.0} \\
        PCNN & 81.9 & 19.6 & \textbf{88.7} & 11.2 & 87.3 & \textbf{73.3} & 80.1 & 26.3 & \textbf{82.7} & 5.01 & 82.1 & \textbf{84.2} & 94.4 & 20.1 & 96.3 & 14.8 & \textbf{97.4} & \textbf{83.8} \\
		\hline
	\end{tabular}}
	\label{ASR of different methods under defenses}
\end{table*}


\subsection{Attack results}
The results of all backdoored models are summarized in Table \ref{ASR of different methods}. It shows that the ASR of our proposed IRBA is competitive with previous baseline backdoor attacks and greater than 90\% on three datasets across all the models, which demonstrates the threat of IRBA in point cloud classification. To verify the stealthiness of IRBA, we further present the ACC of clean-trained models and the backdoored models in Table \ref{acc of clean model} and Table \ref{ASR of different methods}. The average difference of ACC between clean-trained models and IRBA is less than 3\%, which proves that IRBA doesn't drop the ACC of the backdoored model by a large margin. To investigate the influences of different target labels, we evaluate our method with different target labels ($y_t=0, y_t=1, y_t=2$) against PointNet++ in Table \ref{different target label}, which reveals the superior and stable attack performance of IRBA across different target classes.

\subsection{Resistance to various pre-processing techniques}
We evaluate the resistance of three backdoor attacks against the statistical outlier removal (SOR) and random rotation on three datasets across four models. Besides, we further use various transformations in the 3D augmentation to test the robustness of backdoor attacks. The default target model and dataset are PointNet++ and ModelNet10, respectively. \par

\begin{table*}[t]
	\centering
	\caption{ACC (\%) and ASR (\%) of backdoored models with the PointBA-Ball trigger, the PointBA-Rotation trigger, and our IRBA trigger on ModelNet10 under various pre-processing techniques during training. ``R'', ``R3'', ``Scaling'', ``Shift'', ``Dropout'', and ``Jitter'' denote random rotation along the $z$-axis, random rotation along three axes, random scaling, random shift, random dropout, and random jitter for the point clouds, respectively. ``$\checkmark$'' indicates including that technique during the training stage and ``\ding{55}'' indicates not. The best results are highlighted in \bf{bold}.}
    \normalsize
	\setlength{\tabcolsep}{3.2mm}{
	\begin{tabular}{ccccccc|cccccc}
		\hline
    \multicolumn{7}{c|}{Pre-processing Technique} & \multicolumn{6}{c}{Method}  \\ 
    \cline{1-13} 
    \multicolumn{1}{c}{\multirow{2}{*}{SOR}} & \multicolumn{1}{c}{\multirow{2}{*}{R}} & \multicolumn{1}{c}{\multirow{2}{*}{R3}} & \multicolumn{1}{c}{\multirow{2}{*}{Scaling}} & \multicolumn{1}{c}{\multirow{2}{*}{Shift}} & \multicolumn{1}{c}{\multirow{2}{*}{Dropout}} & \multicolumn{1}{c|}{\multirow{2}{*}{Jitter}} & \multicolumn{2}{c}{PointBA-Ball} & \multicolumn{2}{c}{PointBA-Rotation} & \multicolumn{2}{c}{IRBA (Ours)} \\ 
    & & & & & & & ACC & ASR & ACC & ASR & ACC & ASR \\
    \hline
    \ding{55} & \ding{55} & \ding{55} & \ding{55} & \ding{55} & \ding{55} & \ding{55} & 93.2 & \textbf{100} & 92.3 & 98.3 & \textbf{93.2} & 98.1 \\
    \checkmark & \ding{55} & \ding{55} & \ding{55} & \ding{55} & \ding{55} & \ding{55} & 84.4 & 17.2 & 91.3 & \textbf{98.6} & \textbf{92.9} & 97.2 \\
    \checkmark & \checkmark & \ding{55} & \ding{55} & \ding{55} & \ding{55} & \ding{55} & 83.1 & 11.7 & 88.8 & 8.41 & \textbf{92.4} & \textbf{96.6} \\ 
    \checkmark & \checkmark & \checkmark & \ding{55} & \ding{55} & \ding{55} & \ding{55} & 85.7 & 28.9 & 87.5 & 13.7 & \textbf{89.9} & \textbf{72.4} \\
    \checkmark & \checkmark & \checkmark & \checkmark & \ding{55} & \ding{55} & \ding{55} & 90.8 & 6.97 & 90.6 & 6.49 & \textbf{91.2} & \textbf{63.1}  \\ 
    \checkmark & \checkmark & \checkmark & \checkmark & \checkmark & \ding{55} & \ding{55} & 89.7 & 3.96 & 89.2 & 9.86 & \textbf{90.2} & \textbf{60.4} \\ 
    \checkmark & \checkmark & \checkmark & \checkmark & \checkmark & \checkmark & \ding{55} & 90.2 & 4.47 & \textbf{90.9} & 10.3 & 90.4 & \textbf{52.1} \\
    \checkmark & \checkmark & \checkmark & \checkmark & \checkmark & \checkmark & \checkmark & 88.8 & 9.25 & 90.7 & 9.13 & \textbf{90.7} & \textbf{47.5} \\
		\hline
	\end{tabular}}
	\label{ASR of different methods under more augmentations}
\end{table*}

\begin{figure*}[t]
    \begin{minipage}{\textwidth}
    \centering
    \includegraphics[width=0.9\textwidth]{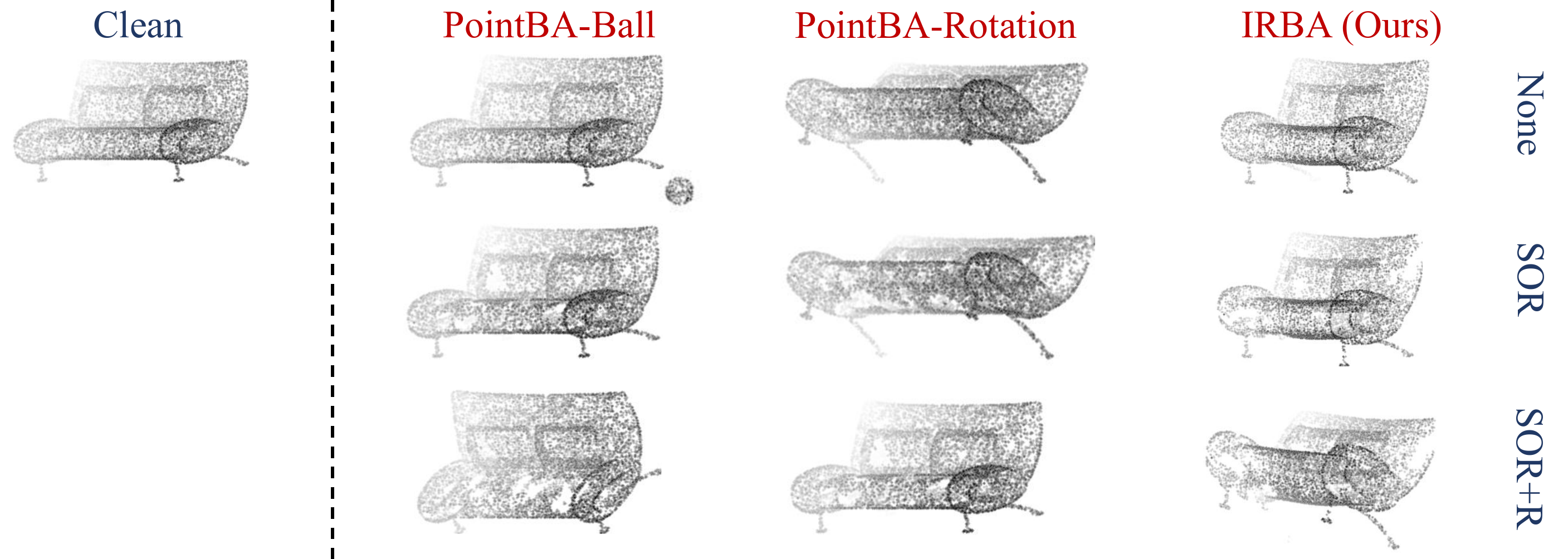}
    \caption{Visualization of the point cloud with the PointBA-Ball trigger \cite{xiang2021backdoor, li2021pointba}, the PointBA-Rotation trigger \cite{li2021pointba} and our IRBA trigger and those after SOR and random rotation.}
	\label{fig:vision sor and rotation}
    \end{minipage}
\end{figure*}

\begin{figure*}[t]
  \begin{minipage}{\linewidth}
    \centering
    \subfloat[PointBA-Ball] {
    \label{interaction trigger}   
    \includegraphics[width=0.27\columnwidth]{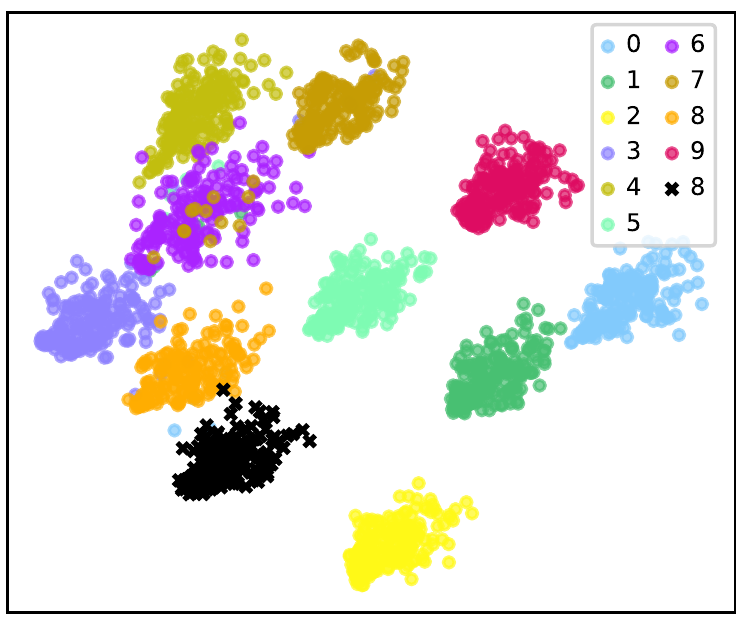}  
    }     
    \subfloat[PointBA-Rotation] { 
    \label{orientation trigger} 
    \includegraphics[width=0.27\columnwidth]{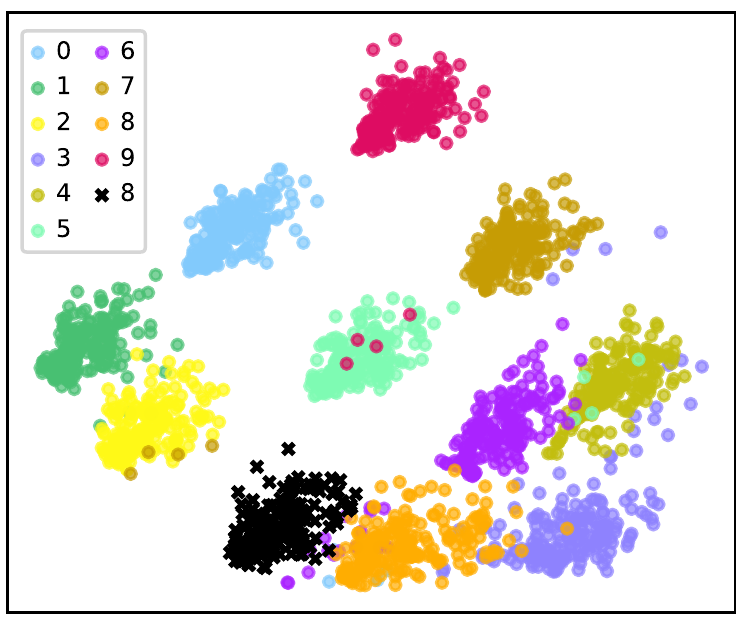}  
    }
    \subfloat[IRBA (Ours)] { 
    \label{irba trigger} 
    \includegraphics[width=0.27\columnwidth]{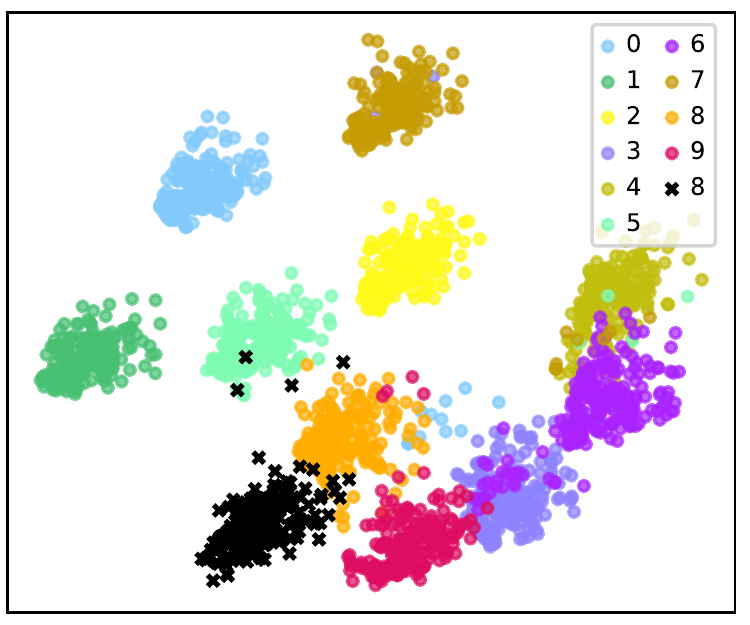} 
    }    
    \caption{t-SNE visualization of the poisoned dataset in the feature space generated by three backdoor attacks on ModelNet10. The target label is chosen as ``Table'' ($y_t=8$), visualized in the black color. }     
    \label{fig:tsne without sor}  
  \end{minipage}
  \begin{minipage}{\linewidth}
    \centering
\subfloat[PointBA-Ball] {
\label{interaction trigger}     
\includegraphics[width=0.27\linewidth]{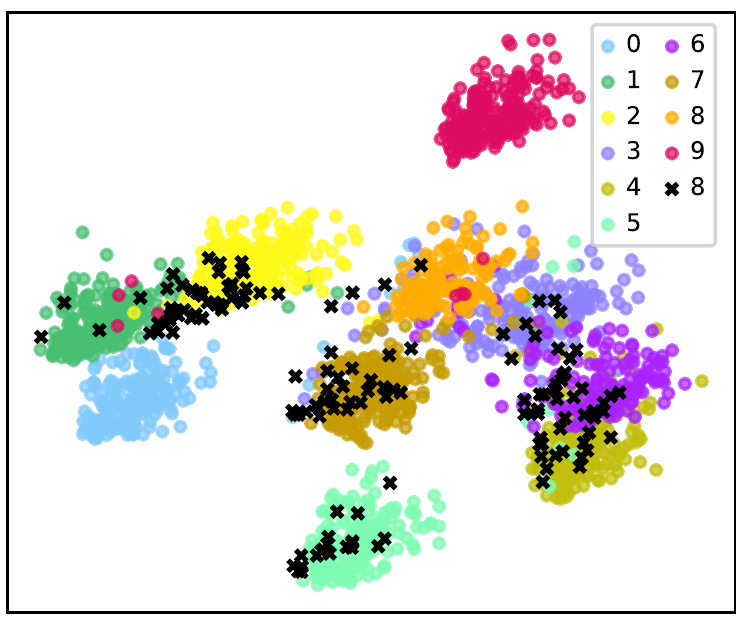}  
}     
\subfloat[PointBA-Rotation] {   
\label{orientation trigger} 
\includegraphics[width=0.27\linewidth]{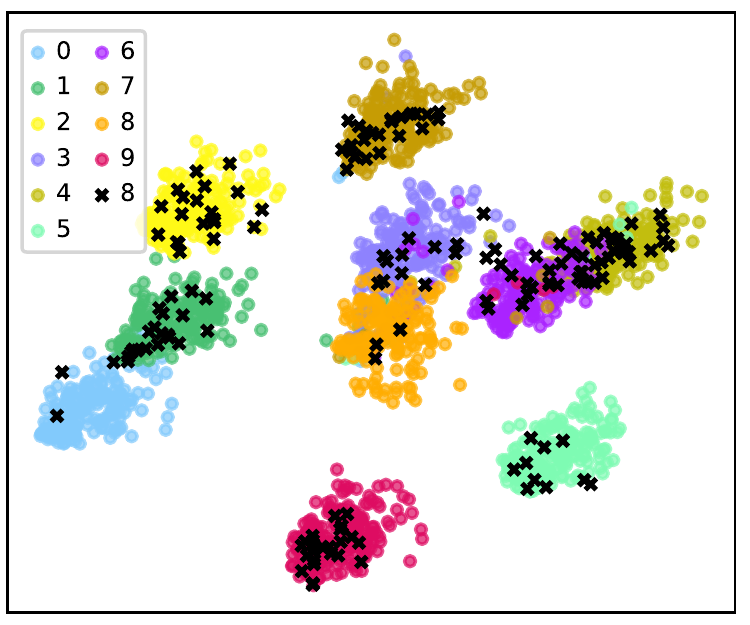}  
}    
\subfloat[IRBA (Ours)] {   
\label{irba trigger} 
\includegraphics[width=0.27\linewidth]{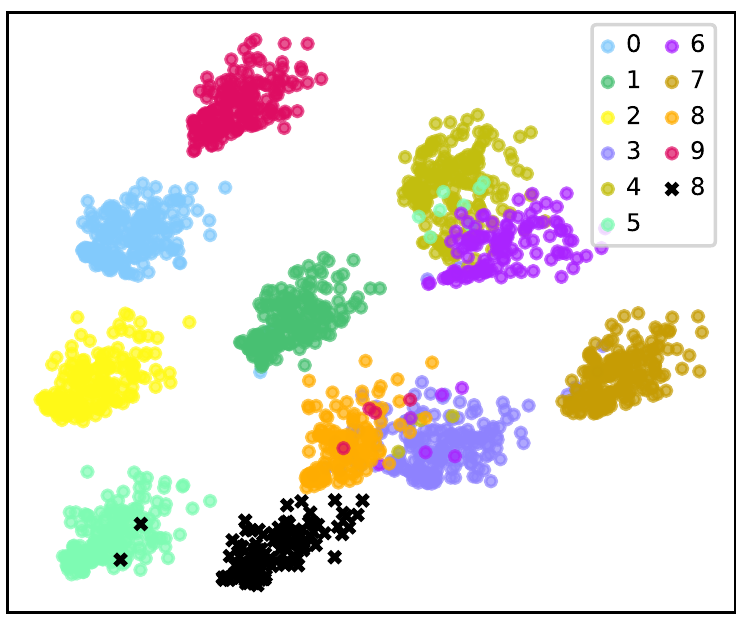} 
}    
\caption{t-SNE visualization of the poisoned dataset in the feature space generated by three backdoor attacks under SOR and random rotation on ModelNet10. The target label is chosen as ``Table'' ($y_t=8$), visualized in the black color. }     
\label{fig:tsne}  
  \end{minipage}
\end{figure*}

\begin{table}[t]
\centering
    \begin{minipage}[t]{\hsize}\centering
    	\caption{ACC (\%) / ASR (\%) of three attacks on ModelNet10 against PointNet++ after SRS and DUP-Net. 
    	}
    	\resizebox{\linewidth}{!}
    	{
 	\begin{tabular}{l|cccccc}
		\hline
		\multicolumn{1}{l|}{\multirow{2}{*}{Pre-processing Technique}} & \multicolumn{2}{c}{PointBA-Ball} & \multicolumn{2}{c}{PointBA-Rotation} & \multicolumn{2}{c}{IRBA}  \\ 
		& ACC & ASR & ACC & ASR & ACC & ASR \\
        \hline
		SRS & 92.9 & 100 & 93.1 & 97.3 &	93.6 & 96.7  \\
		SRS+SOR+R & 81.4 & 38.3 & 87.8 & 10.0 &	91.2 & 81.1  \\
		DUP-Net & 80.6 & 14.4 & 90.4 & 98.7 &	93.5 & 98.3   \\
		DUP-Net+SOR+R & 81.1 & 16.3 & 82.8 & 39.5 & 93.3 & 98.1   \\
		\hline
	\end{tabular}}
    	\label{SRS and DUP-Net}
    \end{minipage}
\end{table}

\begin{table*}[t]
\centering
\makebox[0pt][c]{\parbox{\textwidth}{%
    \begin{minipage}[t]{0.49\hsize}\centering
    	\caption{Scores of the human perceptual study. A lower score corresponds to more imperceptible poisoned samples.}
    	\resizebox{0.95\linewidth}{!}
    	{
    	\begin{tabular}{l|ccc}
    		\hline
    		Dataset & PointBA-Ball & PointBA-Rotation & IRBA (Ours)  \\ \cline{1-4}
    		ModelNet10 & 4.8 &	2.8 & \textbf{2.2}  \\
    		ModelNet40 & 4.9 &	\textbf{1.9} &	2.4  \\
    		ShapeNetPart & 4.9 & 1.7 & \textbf{1.4}  \\
    		\hline
    	\end{tabular}}
    	\label{human percep}
    \end{minipage}
    \hfill
    \begin{minipage}[t]{0.49\hsize}\centering
    	\caption{CD $\times$ 100 of existing methods and IRBA on three datasets. The lower is better.}
    	\resizebox{0.95\linewidth}{!}
    	{
    	\begin{tabular}{l|ccc}
    		\hline
    		Dataset & PointBA-Ball & PointBA-Rotation & IRBA (Ours)  \\ \cline{1-4}
    		ModelNet10 & 1.16 &	\textbf{0.12} & 0.47  \\
    		ModelNet40 & 1.14 &	\textbf{0.13} &	0.52  \\
    		ShapeNetPart & 0.86 & 0.14 & \textbf{0.12}  \\
    		\hline
    	\end{tabular}}
    	\label{cd evaluation}
    \end{minipage}
}}
\end{table*}

\noindent \textbf{Resistance to SOR and random rotation.} Zhou \etal \cite{zhou2019dup} proposed the statistical outlier removal (SOR) to resist point addition attacks by removing points far from the neighbors. We combine the SOR and random rotation along the $z$-axis as the data pre-processing techniques during the training stage. For SOR, we calculate the distance between every point and their 30 nearest points and then remove the 100 points with the farthest neighbor distance. As for the random rotation, the random angle is under $20^\circ$. Table \ref{ASR of different methods under defenses} demonstrates that the SOR and the random rotation can degrade the PointBA-Ball and PointBA-Rotation backdoor attack performance by a large margin but have little effect on IRBA. To explore why our method can still achieve a high ASR compared with the other two backdoor attacks, we visualize samples, the feature space of the poisoned dataset and that after SOR and random rotation in Fig. \ref{fig:vision sor and rotation}, Fig. \ref{fig:tsne without sor} and Fig. \ref{fig:tsne}. 

As shown in Fig. \ref{fig:vision sor and rotation}, the PointBA-Ball trigger can be completely eliminated after SOR. Although SOR doesn't break the PointBA-Rotation trigger, it can be undermined easily by the random rotation. Hence, due to the damage of the trigger of two attacks, the model is difficult to distinguish the features between clean samples and poisoned samples, which can be seen in Fig. \ref{fig:tsne}. As for IRBA, it does not introduce outliers such that it is robust to SOR
and the linear operation rotation is not enough to destruct the unique structure generated by our nonlinear method. Fig. \ref{fig:tsne} shows that our poisoned samples can cluster independently from other classes in the feature space, indicating that our attack succeeds to implant the backdoor into the model.\par

\noindent \textbf{Resistance to various augmentations.} In addition to SOR and the random rotation, we further apply five data augmentations to explore the effect on three attacks, including random rotation under $360^\circ$ along three axes, random scaling from 0.5 to 1.5, random shift ranging from -0.1 to 0.1, random dropout with the max ratio 0.2, random jitter in a Gaussian distribution $\mathcal{N}(0, 0.02)$, and clipping within 0.05. As shown in Table \ref{ASR of different methods under more augmentations}, our method can achieve a 
higher ASR than the PointBA-Ball and PointBA-Rotation backdoor attack under all the above augmentations during the backdoor training. Although ASR of IRBA decreases with the accumulation of various augmentations, IRBA can still achieve at least 47.5\% ASR, which benefits from the nonlinear and local distortion between our poisoned samples and clean ones.

\noindent \textbf{More pre-processing techniques}
We try the defaulted Simple Random Sampling (SRS) \cite{yang2019adversarial} during the training and show the results in Table \ref{SRS and DUP-Net}. Besides, we utilize the original upsampling network in DUP-Net \cite{zhou2019dup} to upsample the training dataset from 1024 points to 2048 points after SOR layer and the results are shown in Table \ref{SRS and DUP-Net}. All the above results demonstrate that IRBA can resist these two pre-processing techniques, which verifies its superiority. \par

\begin{figure*}[h]
  \begin{minipage}{\linewidth}
    \centering
\subfloat[PointBA-Ball] {
\label{interaction trigger}   
\includegraphics[width=0.28\columnwidth]{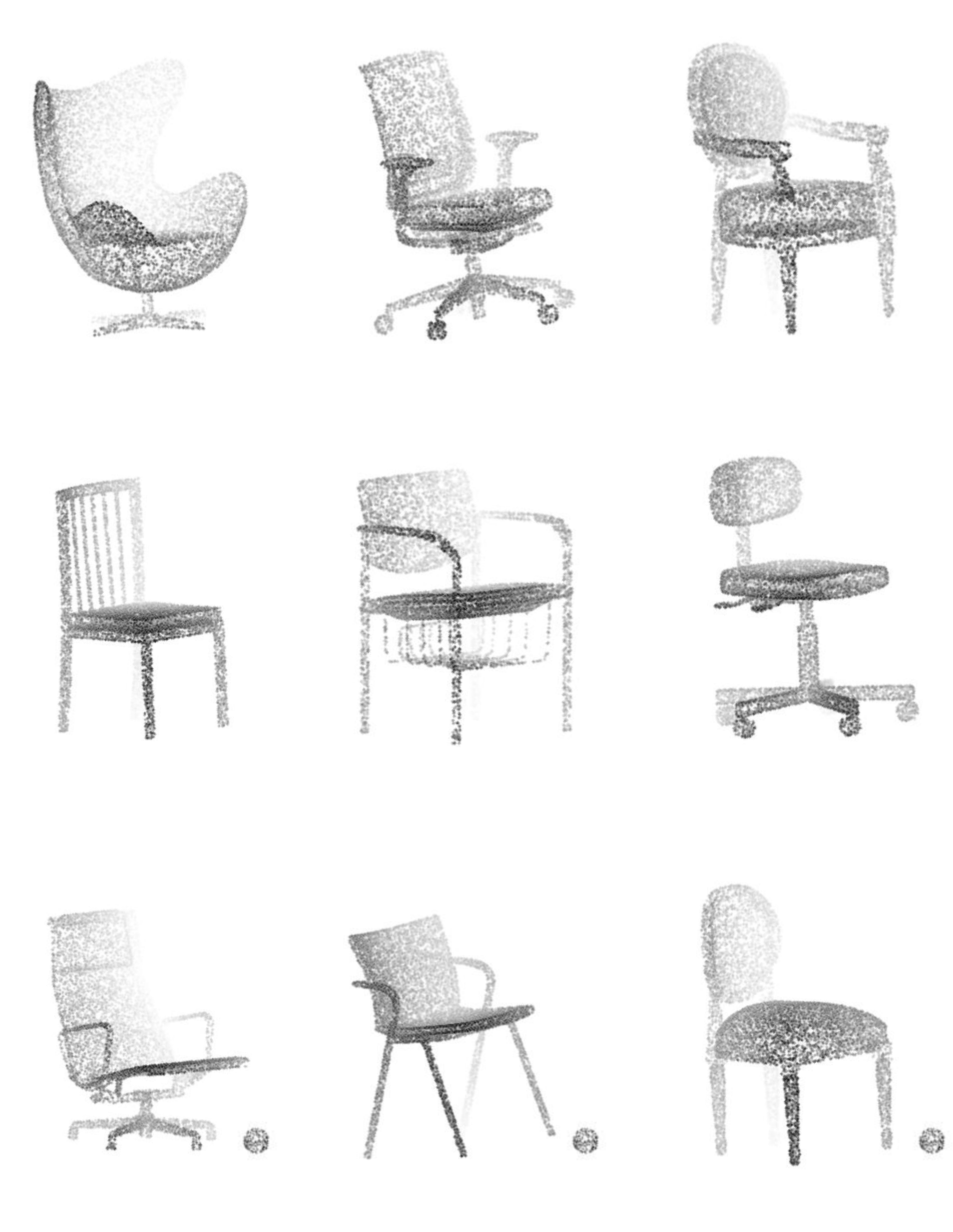}  
}     
\hspace{1.5em}
\subfloat[PointBA-Rotation] { 
\label{orientation trigger} 
\includegraphics[width=0.29\columnwidth]{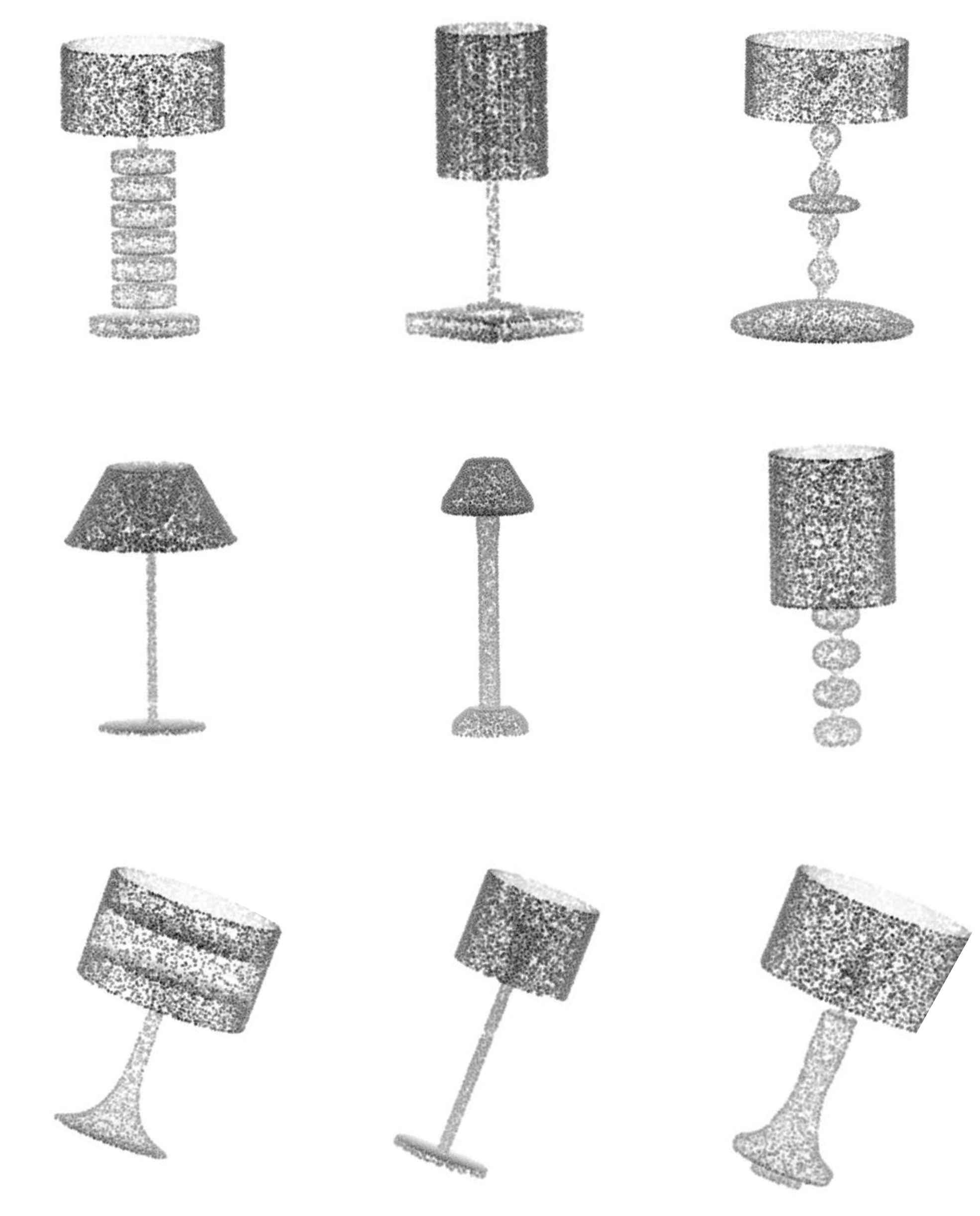}     
}
\hspace{1.5em}
\subfloat[IRBA (Ours)] { 
\label{irba trigger} 
\includegraphics[width=0.28\columnwidth]{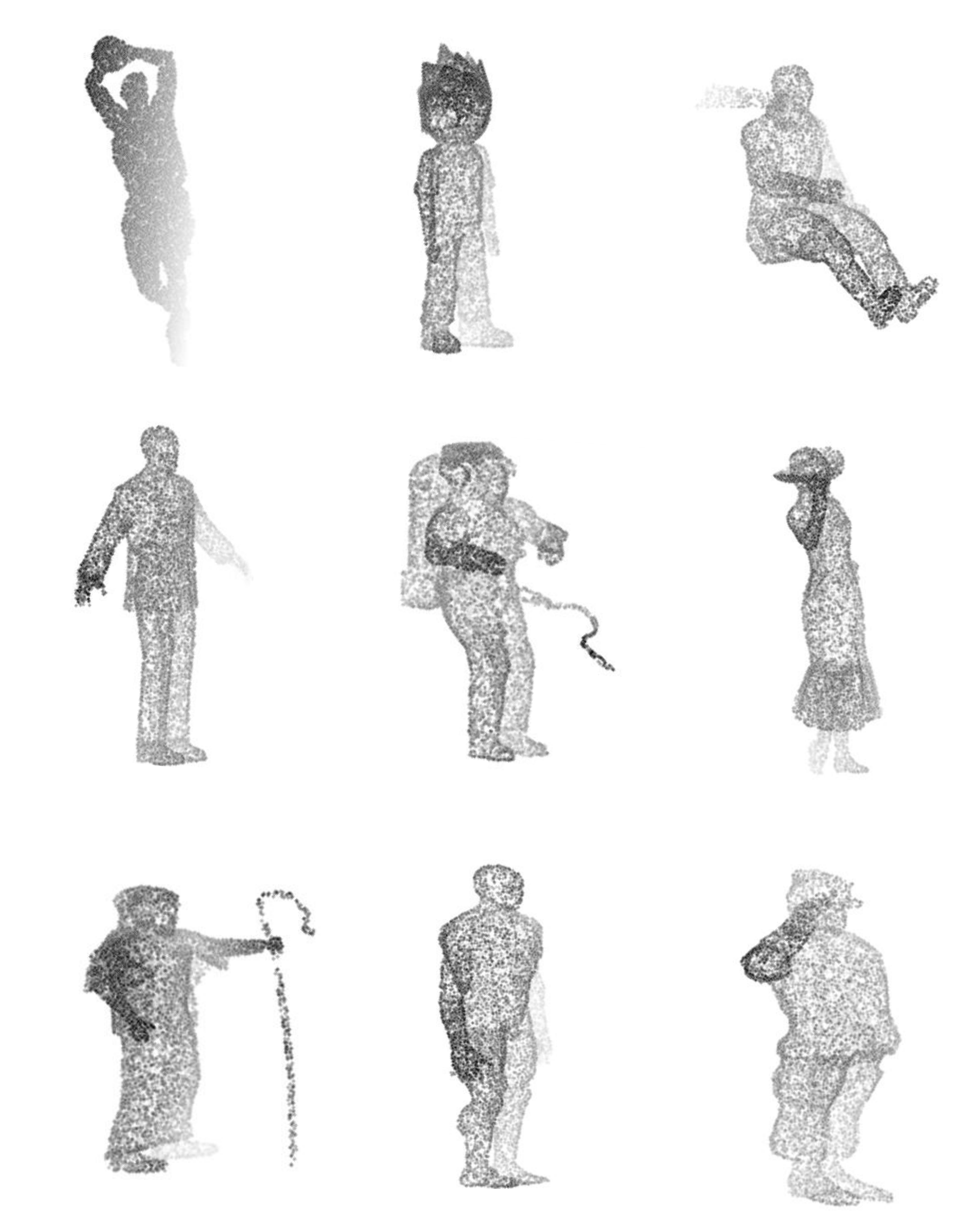} 
}    
\caption{The randomly selected sample sets of three attacks used in human perceptual study. }     
\label{fig:human study}  
  \end{minipage}

\begin{minipage}{\textwidth}
\centering
\includegraphics[width=0.9\textwidth]{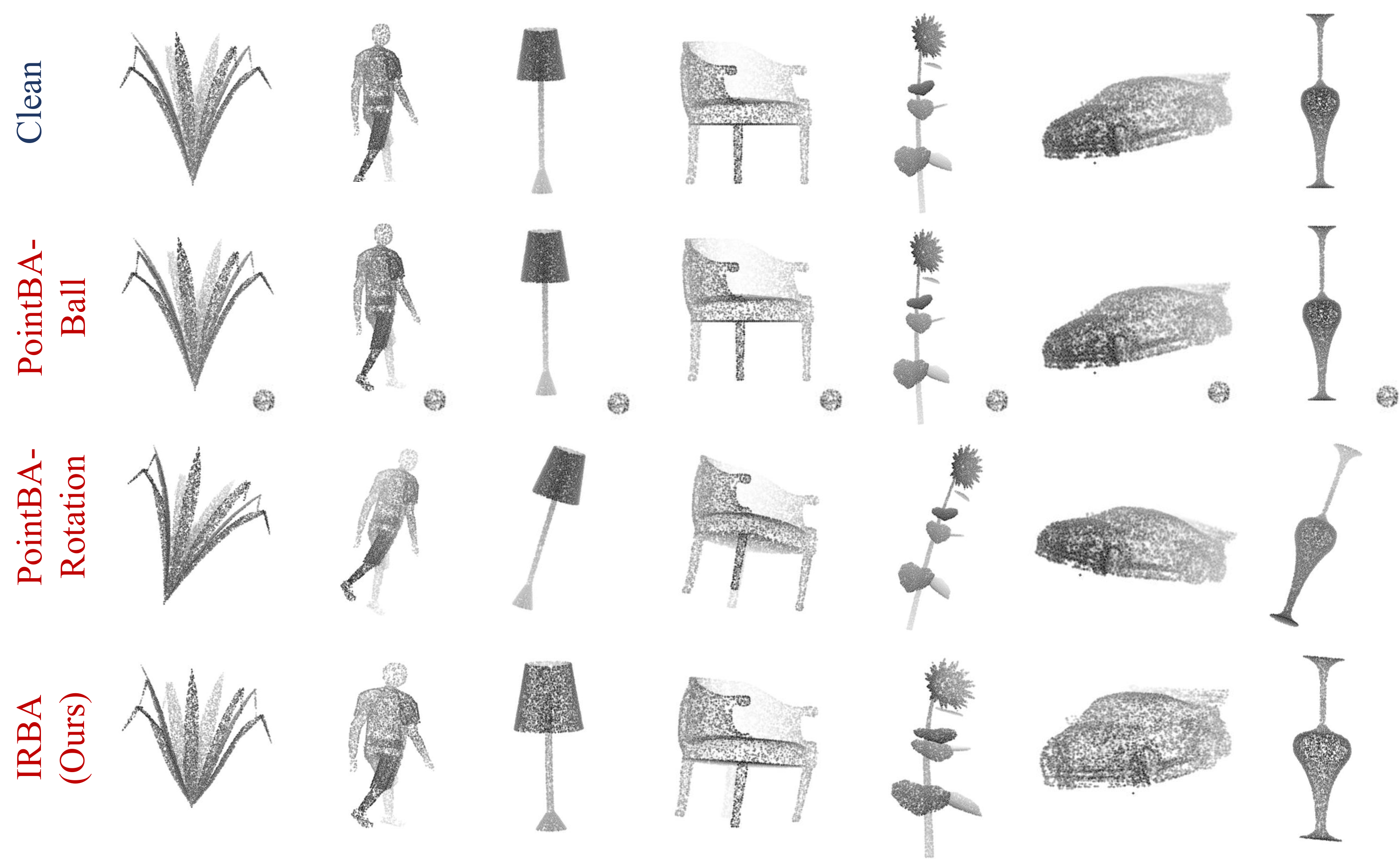}
\caption{Visualizations of poisoned samples generated by the PointBA-Ball attack \cite{xiang2021backdoor, li2021pointba}, the PointBA-Rotation attack \cite{li2021pointba}, and our IRBA.}
\label{fig:vision all example}
\end{minipage}
  
\end{figure*}

\subsection{Human perceptual study}
\label{Human Perceptual Study}

To evaluate the visual perceptibility of poisoned samples generated by different attack methods, we conduct a human perceptual study in this section. We list overall results in Table \ref{human percep}. In our study, we randomly mix clean samples and poisoned samples and show them to 20 participants. These participants are asked to give a score $\in \{1, 2, 3, 4, 5\}$ to judge the probability that some of the samples have been modified. A lower score corresponds to a lower probability, \ie, a less perceptible poisoned sample. \par

We can observe that the score of the samples with the PointBA-Ball trigger is the highest because it contains noticeable additional points. In contrast, the poisoned samples of the PointBA-Rotation method and IRBA achieve relatively low scores, which indicates the visual stealthiness of these two backdoor attacks. To quantify the imperceptibility of three backdoor attacks, we also calculate the CD $\times$ 100 on three datasets, as shown in Table \ref{cd evaluation}. It indicates that the PointBA-Rotation attack and our IRBA can achieve the lowest CD on ModelNet and ShapeNet, respectively. 
The above results and visualizations confirm that IRBA is hardly perceptible and is difficult to be spotted by humans when our poisoned samples are mixed with clean ones.
We randomly select a sample set used in our human perceptual study for each attack and visualize them in Fig. \ref{fig:human study}. The first two rows are clean samples and the third row is the poisoned samples. Besides, more poisoned samples generated by the the PointBA-Ball attack, the PointBA-Rotation attack, and IRBA are shown in Fig. \ref{fig:vision all example}.

\begin{table*}[ht]
	\caption{ACC (\%) and ASR (\%) of backdoored models with our IRBA trigger on ModelNet10 under adaptive defense during training.}
	\label{adaptive augmentation}
    \normalsize
	\centering
	\setlength{\tabcolsep}{4mm}{
	\begin{tabular}{ccc|cccc}
		\hline
    \multicolumn{1}{c}{\multirow{2}{*}{Range of $\alpha$}} &  \multicolumn{1}{c}{\multirow{2}{*}{Range of $s$}} & \multicolumn{1}{c|}{\multirow{2}{*}{Range of $W$}} & \multicolumn{2}{c}{Average} & \multicolumn{2}{c}{Smooth}  \\
    & & & ACC & ASR & ACC & ASR \\ \hline
    -10 $\sim$ 10 & 1 $\sim$ 10 & 1 $\sim$ 32 & 93.8 & 93.4 & 93.1 & 75.9 \\
    -10 $\sim$ 10 & 1 $\sim$ 10 & 1 $\sim$ 16 & 93.3 & 89.2 & 92.9 & 65.6 \\
    -10 $\sim$ 10 & 1 $\sim$ 10 & 1 $\sim$ 8 & 94.2 & 90.9 & 92.8 & 55.4 \\
    -10 $\sim$ 10 & 1 $\sim$ 10 & 1 $\sim$ 4 & 92.9 & 89.4 & 91.7 & 60.3 \\
    -10 $\sim$ 10 & 1 $\sim$ 10 & 1 $\sim$ 2 & 92.5 & 90.7 & 91.7 & 79.8 \\
    -5 $\sim$ 5 & 1 $\sim$ 5 & 1 $\sim$ 32 & 93.5 & 98.1 & 93.3 & 88.7 \\ 
    -5 $\sim$ 5 & 1 $\sim$ 5 & 1 $\sim$ 16 & 92.7 & 96.9 & 92.9 & 90.1 \\ 
    -5 $\sim$ 5 & 1 $\sim$ 5 & 1 $\sim$ 8 & 93.2 & 94.5 & 93.2 & 90.9 \\ 
    -5 $\sim$ 5 & 1 $\sim$ 5 & 1 $\sim$ 4 & 93.8 & 93.3 & 92.2 & 81.3 \\ 
    -5 $\sim$ 5 & 1 $\sim$ 5 & 1 $\sim$ 2 & 93.5 & 92.7 & 92.0 & 90.9 \\ 
	    \hline
	\end{tabular}
	}
\end{table*}

\begin{figure*}[t]
	\centering
	\begin{minipage}{0.45\linewidth}
        \includegraphics[width=\linewidth]{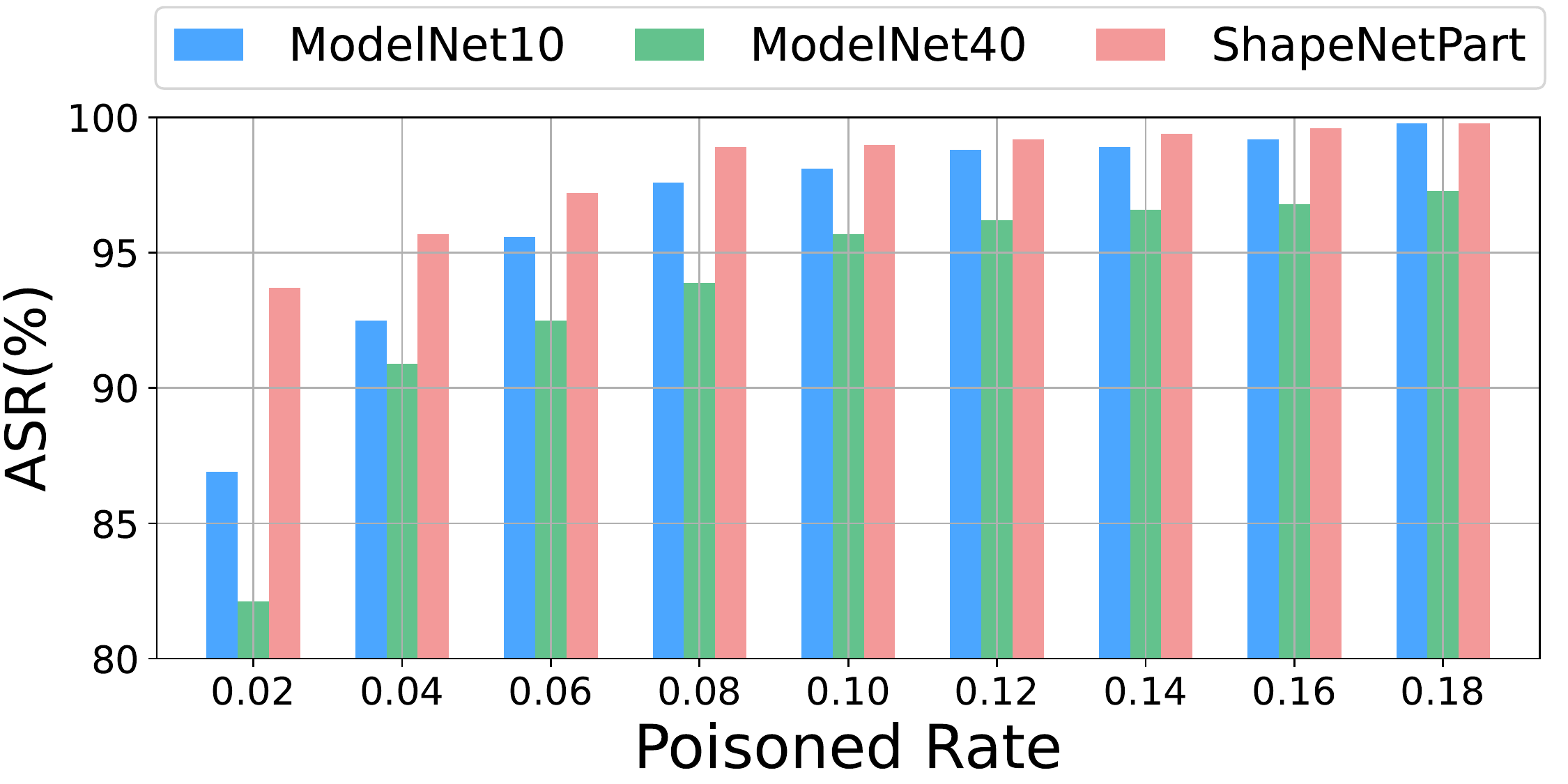}
            \vspace{-0.5em}
        	\caption{ASR (\%) of IRBA with different poisoned rates on three datasets.}
        \label{fig:poisoned rate}   
    \end{minipage}
    \hfill
    \begin{minipage}{0.49\linewidth}
        \begin{minipage}{0.43\linewidth}
        \includegraphics[width=\linewidth]{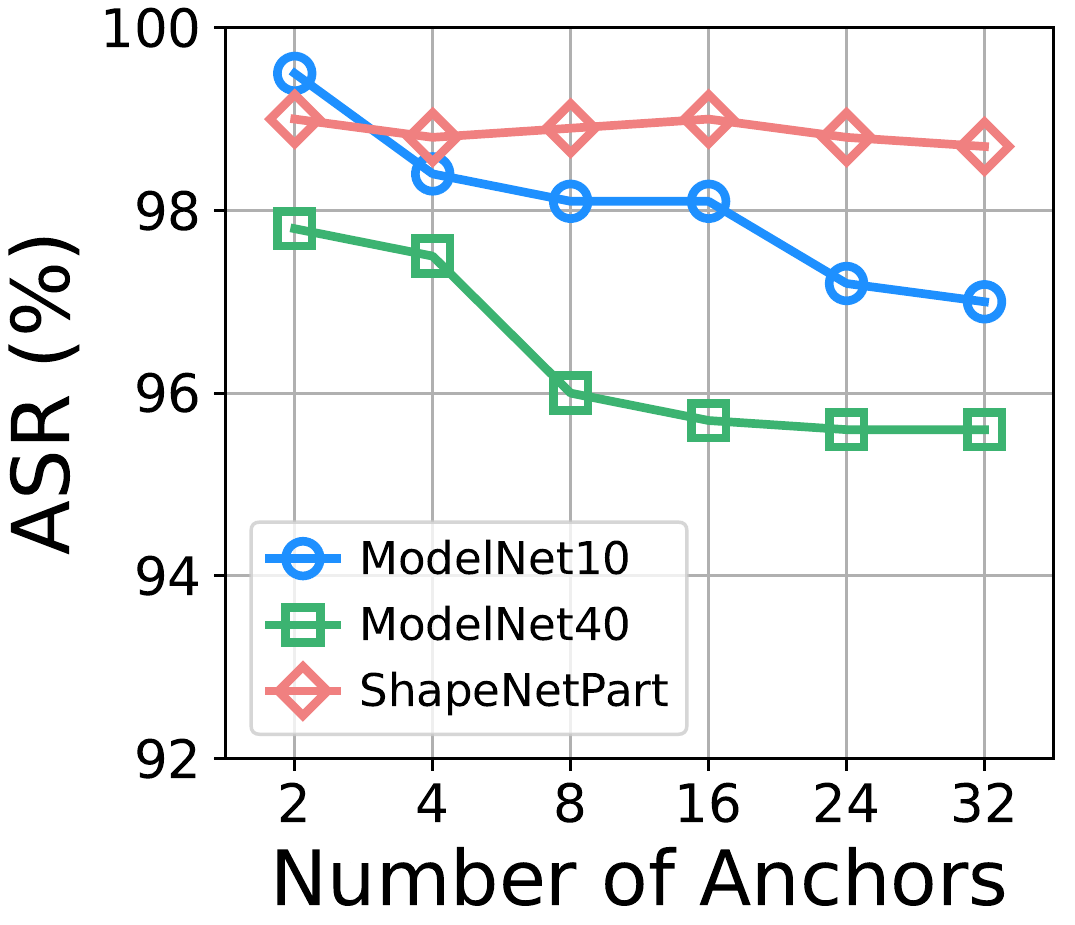} 
        \end{minipage}
    \hspace{-0.6em}
        \begin{minipage}{0.51\linewidth}
        \includegraphics[width=\linewidth]{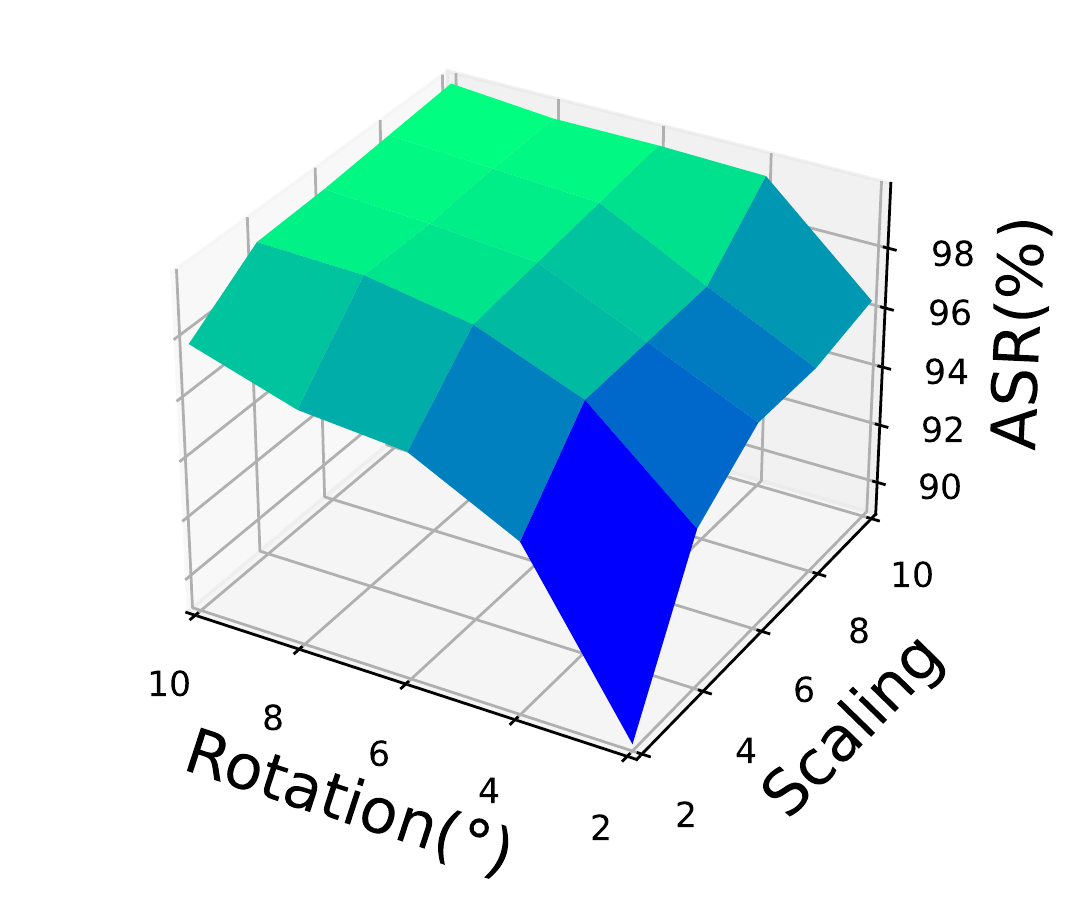}
        \end{minipage}
        \vspace{-0.5em}
    	\caption{Left: ASR (\%) of IRBA with different numbers of anchors $W$ on three datasets. Right: ASR (\%) of IRBA with different rotation angles $\alpha$ and scaling sizes $s$ on ModelNet10.}
        \label{fig:trigger parameter}
    \end{minipage}
\end{figure*}

\begin{figure}[t]
    \centering
    \begin{minipage}{\linewidth}
        \begin{minipage}{0.48\linewidth}
        \includegraphics[width=\linewidth]{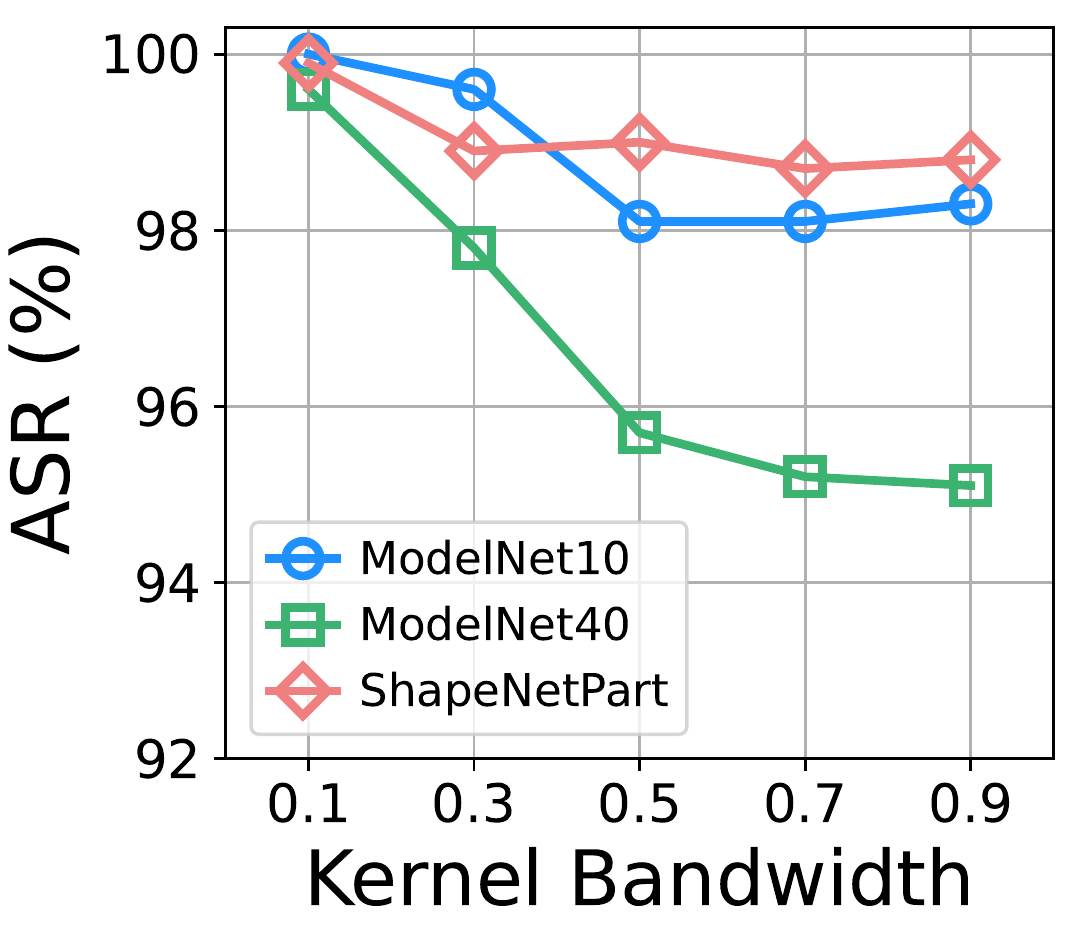} 
        \end{minipage}
        \begin{minipage}{0.48\linewidth}
        \includegraphics[width=\linewidth]{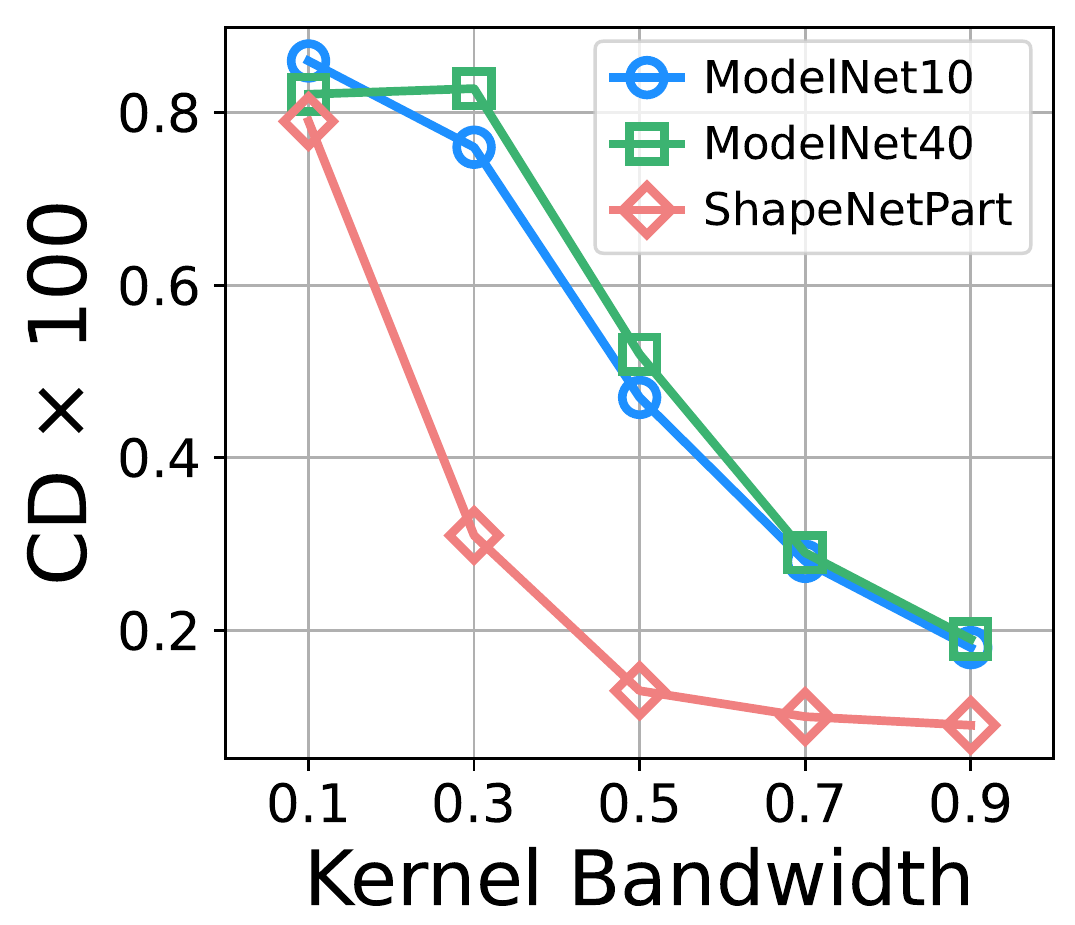}
        \end{minipage}
    	\caption{ASR (\%) and CD $\times$ 100 of IRBA with different kernel bandwidths $h$ on three datasets.}
        \label{fig:kernel bandwidth}
    \end{minipage}
\end{figure}

\subsection{Ablation study}
\noindent \textbf{Effect of the poisoned rate.} We explore the effect of the poisoned rate towards our proposed attack on three datasets and show the results in Fig. \ref{fig:poisoned rate}. It can be seen that the attack effectiveness becomes stronger with the increase of the poisoned rate when it is lower than 0.08. Meanwhile, once the poisoned rate is larger than 0.1, the ASR of our attack can exceed 95\% and remain at a high level across three datasets. The high ASR with a low poisoned rate can ensure the stealthiness of IRBA.

\noindent \textbf{Effect of the number of anchors $W$.} In this part, we sample different numbers of the anchors $W$ to investigate the effect on our attack. From the results in Fig. \ref{fig:trigger parameter}, we can observe that fewer anchors will lead to a higher ASR on three datasets. As such, the anchor-based transformation will be performed within a more finite neighborhood of the chosen anchors and result in an irregular deformation on the shape of the point cloud. The visualization and CD score are shown in Fig. \ref{fig:vis_w_parameter}.

\noindent \textbf{Effect of the rotation angle $\alpha$ and scaling size $s$.} Here, we present the results under various rotation angles $\alpha$ and scaling sizes $s$ of our trigger. As shown in Fig. \ref{fig:trigger parameter}, both $\alpha$ and $s$ can affect the ASR of our attack jointly, where the larger $\alpha$ and $s$ indicate the higher attack performance. Our attack can achieve 90\% ASR even though we apply a weak rotation and scaling as the trigger injection, \ie, $\alpha=2^\circ$ and $s=2$. Specially, when the value of $\alpha$ and $s$ outperforms 8, the ASR of our attack can reach above 99\%. However, we should note that the larger $\alpha$ and $s$ will cause more severe deformation of the original structure. Therefore, it is necessary to choose the appropriate rotation angle and scaling size to balance the ASR and imperceptibility of the attack.

\noindent \textbf{Effect of the kernel bandwidth.}
We explore the effect of the kernel bandwidth $h$ in the smooth aggregation on our attack. The ASR and CD score are shown in Fig. \ref{fig:kernel bandwidth}. We can see that the ASR drops with the increase of the kernel bandwidth $h$. This is because the difference between the clean sample and the poisoned sample (\textit{i.e.}, the CD score) is lower and the poisoned samples become more difficult to be detected when the kernel bandwidth $h$ increases.

\subsection{Potential adaptive defense}
\label{sec: adpative defense}
In the above experiments, we assume that the victim has no information about our attack. In this section, we consider a more challenging setting, where the victim knows the existence of IRBA and trains the model with an adaptive defense. Based on the victim's knowledge, the adaptive defense can be divided into two categories. One is that the victim only knows the multi-anchor transformation but aggregates the transformed point clouds in an average weight manner. The other is that both multi-anchor transformation and smooth aggregation are known by the victim. However, the victim can not get the specific parameters of IRBA in both settings. We denote them as ``Average'' and ``Smooth'', respectively. We keep the settings of our IRBA unchanged as Section \ref{sec:setting} and vary the defense parameters. The results are shown in Table \ref{adaptive augmentation}. \par

We can see that IRBA can still achieve about 90\% ASR under ``Average'' defense for different parameters, which shows IRBA can resist it successfully. This is because the nonlinear operation in smooth aggregation can generate more diverse 3D shapes to make IRBA robust against the linear average aggregation. As for the smooth adaptive defense, we set the range of the bandwidth $h$ from 0.1 to 0.9. Table \ref{adaptive augmentation} shows that despite the ``Smooth'' defense during backdoor training, the ASR of IRBA is still greater than 55.4\%. It may be because that there are many parameters in IRBA including the rotation angle, scaling size, smoothing parameter, \etc. Although the victim adopts smooth aggregation, the effect of the poisoned samples remains, due to the unknown parameters in our poisoned samples generation. 



\section{Conclusion and social impacts}
\label{sec:Conclusion and Social Impacts}
In this paper, we propose the imperceptible and robust backdoor attack (IRBA) in 3D domain. IRBA transforms the point cloud centered at multiple sampled anchors and smoothly aggregates them into one poisoned sample. The dataset with a small fraction of these poisoned samples can successfully inject the backdoor behavior into the point-based model trained on it. Extensive experiments demonstrate that our backdoor not only bypasses most pre-processing techniques in point cloud  but also is imperceptible to human inspection. We hope that the proposed IRBA can serve as an effective tool to examine the vulnerability of 3D models.\par

As for the social impacts, the main threat of IRBA can happen in the autonomous driving applications. Suppose that the autonomous vehicle manufacturers outsource the dataset to be used and the dataset is unfortunately poisoned with a fraction of samples with the trigger by the attacker. Once the autonomous vehicle manufacturers deploy their models trained on the malicious dataset, the attacker can adopt the 3D printing technique or LIDAR-based spoofing attacks \cite{cao2019adversarial,sun2020towards} to mislead the autonomous driving model to make a wrong decision, which leads to serious traffic accidents. \par

Please note that we restrict all experiments in the laboratory environment and do not support our attack in the real scenario. The purpose of our work is to raise the awareness of the security concern in 3D community and call for 3D deep learning practitioners to pay more attention to the training data integrity and model trustworthy deployment.

\bibliographystyle{plain}
\bibliography{egbib}

\end{document}